\documentclass{article}

\usepackage[nonatbib,preprint]{neurips_data_2023}

\usepackage[utf8]{inputenc} %
\usepackage[T1]{fontenc}    %
\usepackage[table,xcdraw]{xcolor}         %
\definecolor{citecolor}{HTML}{2980b9}
\definecolor{linkcolor}{HTML}{c0392b}
\usepackage{url}            %
\usepackage{booktabs}       %
\usepackage{amsfonts}       %
\usepackage{nicefrac}       %
\usepackage{wrapfig}
\usepackage{microtype}      %
\usepackage{adjustbox}
\newcommand{\sectioncolor}{citecolor}
\usepackage{mdframed}
\usepackage{makecell}
\usepackage{xspace}
\usepackage{subcaption}
\usepackage{caption}
\usepackage[pagebackref=true,breaklinks=true,colorlinks,bookmarks=false,citecolor=citecolor,linkcolor=linkcolor]{hyperref}

\usepackage{array}
\usepackage[square,numbers]{natbib}
\usepackage{tabularx}
\usepackage[colorinlistoftodos]{todonotes}

\def\dataset{Objaverse-XL\xspace}
\def\datasetone{Objaverse 1.0\xspace}

\usepackage{tikz}

\usepackage{refcount}
\newcounter{pagestoreach}
\title{
\dataset: A Universe of 10M+ 3D Objects
}

\author{%
\textbf{Matt Deitke$^{\dagger\psi}$, Ruoshi Liu$^\gamma$, Matthew Wallingford$^\psi$, Huong Ngo$^\psi$, Oscar Michel$^\dagger$,}\\
\textbf{Aditya Kusupati$^\psi$, Alan Fan$^\psi$, Christian Laforte$^\sigma$, Vikram Voleti$^\sigma$, Samir Yitzhak Gadre$^\gamma$,}\\
\textbf{Eli VanderBilt}$^{\dagger}$, \textbf{Aniruddha Kembhavi$^{\dagger\psi}$, Carl Vondrick$^\gamma$, Georgia Gkioxari$^\delta$,}\\
\textbf{Kiana Ehsani$^\dagger$, $^*$Ludwig Schmidt$^{\dagger\psi\ell}$, $^*$Ali Farhadi$^\psi$}\\[0.05in]
$^\dagger$Allen Institute for AI \quad $^\psi$University of Washington, Seattle \quad
$^\gamma$Columbia University\\
$^\sigma$Stability AI \quad
$^\delta$California Institute of Technology \quad
$^\ell$LAION\\
$^*$\texttt{Equal Senior Contribution}
}

\begin{document}

\maketitle

\begin{figure}[h!]
  \vspace{-0.45in}
  \centering
  \includegraphics[width=\textwidth]{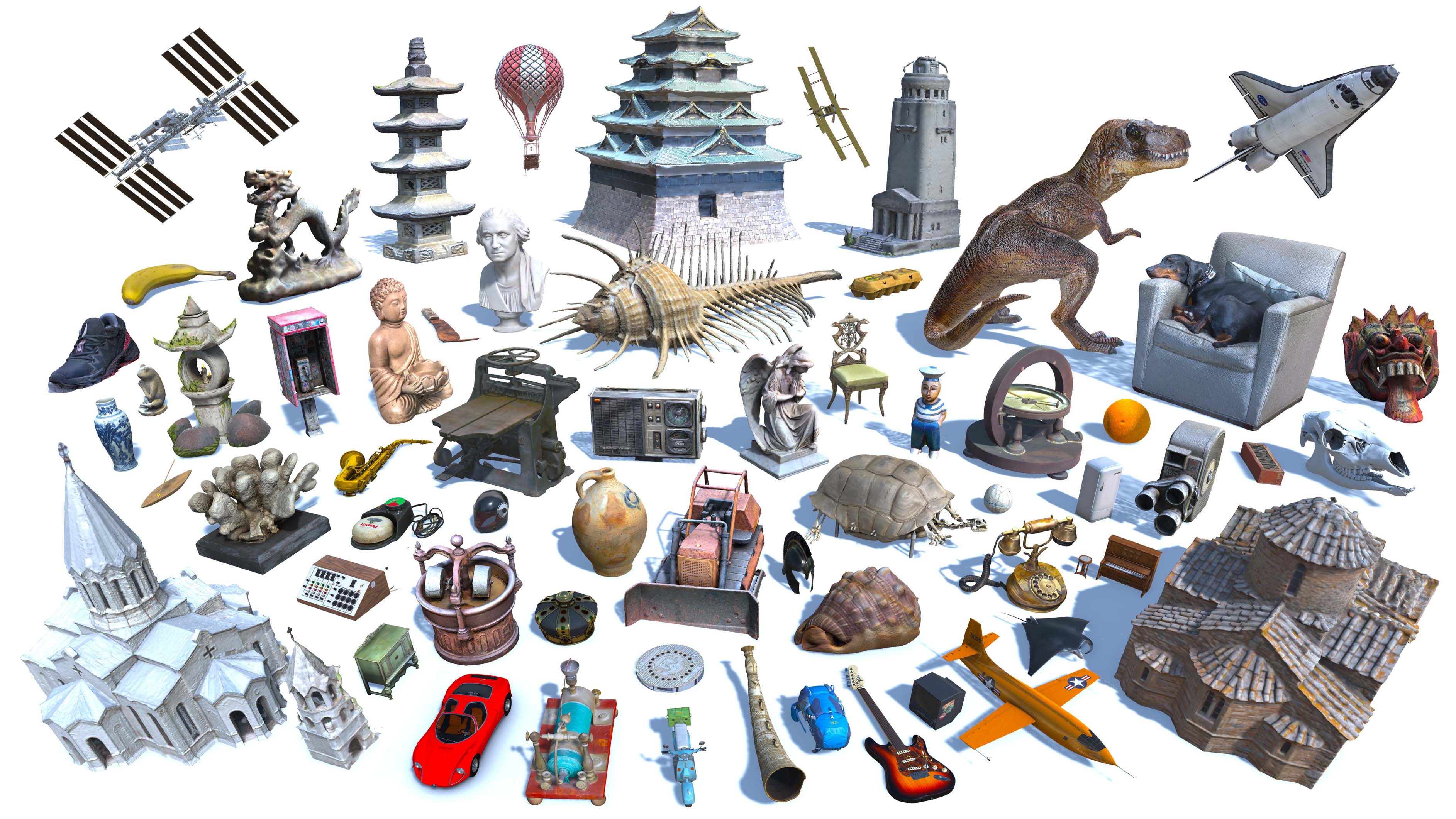}
  \caption{Objaverse-XL includes a ginormous collection of diverse 3D objects from a variety of sources. Here, we show examples of objects in Objaverse-XL rendered in a scene.}
  \vspace{-0.05in}
  \label{fig:teaser}
\end{figure}

\begin{abstract}
Natural language processing and 2D vision models have attained remarkable proficiency on many tasks primarily by escalating the scale of training data. However, 3D vision tasks have not seen the same progress, in part due to the challenges of acquiring high-quality 3D data. In this work, we present Objaverse-XL, a dataset of over 10 million 3D objects. Our dataset comprises deduplicated 3D objects from a diverse set of sources, including manually designed objects, photogrammetry scans of landmarks and everyday items, and professional scans of historic and antique artifacts. Representing the largest scale and diversity in the realm of 3D datasets, Objaverse-XL enables significant new possibilities for 3D vision. Our experiments demonstrate the improvements enabled with the scale provided by Objaverse-XL. We show that by training Zero123 on novel view synthesis, utilizing over 100 million multi-view rendered images, we achieve strong zero-shot generalization abilities. We hope that releasing Objaverse-XL will enable further innovations in the field of 3D vision at scale.
\end{abstract}

\section{Introduction}

Scale has been paramount to recent advances in AI. Large models have produced breakthroughs in language comprehension and generation~\cite{brown2020language,openai2023gpt}, representation learning~\cite{radford2021learning}, multimodal task completion~\cite{alayrac2022flamingo,Lu2022UnifiedIOAU}, image generation~\cite{ramesh2022hierarchical,rombach2022high}, and more. With an increasing number of learnable parameters, modern neural networks consume increasingly large volumes of data. As data has scaled up, the capabilities exhibited by models has dramatically increased.

Just a few years ago, GPT-2~\cite{radford2019language} broke data barriers by consuming roughly 30 billion language tokens and demonstrated promising zero shot results on NLP benchmarks. Now, models like Chinchilla~\cite{hoffmann2022training} and LLaMA~\cite{touvron2023llama} consume trillions of web crawled tokens and easily surpass GPT-2 at benchmarks and capabilities. In computer vision, ImageNet~\cite{deng2009imagenet}, with 1 million images, was the gold standard for representation learning until scaling to billions of images, via web crawled datasets like LAION-5B~\cite{schuhmann2022laion}, produced powerful visual representations like CLIP~\cite{radford2021learning}. Key to scaling up from millions of data points to billions and beyond has been the shift from assembling datasets manually to assembling them from diverse sources via the web.

As language and image data has scaled up, applications that require other forms of data have been left behind. Notable are applications in 3D computer vision, with tasks like 3D object generation and reconstruction, continue to consume small handcrafted datasets. 3D datasets such as ShapeNet~\cite{chang2015shapenet} rely on professional 3D designers using expensive software to create assets, making the process tremendously difficult to crowdsource and scale. The resulting data scarcity has become a bottleneck for learning-driven methods in 3D computer vision. For instance, 3D object generation currently lags far behind 2D image generation, and current 3D generation approaches often still leverage models trained on large 2D datasets instead of being trained on 3D data from scratch. As demand and interest in AR and VR technologies goes up, scaling up 3D data is going to be increasingly crucial.

We introduce Objaverse-XL, a large-scale, web-crawled dataset of 3D assets. Advances in 3D authoring tools, demand, and photogrammetry, have substantially increased the amount of 3D data on the Internet. This data is spread across numerous locations including software hosting services like Github, specialized sites for 3D assets like Sketchfab, 3D printing asset sources like Thingiverse, 3D scanning platforms like Polycam, and specialized sites like the Smithsonian Institute. Objaverse-XL crawls such sources for 3D objects, providing a significantly richer variety and quality of 3D data than previously available, see Figure~\ref{fig:teaser}. Overall, Objaverse-XL comprises of over 10 million 3D objects, representing an order of magnitude more data than the recently proposed Objaverse 1.0~\cite{deitke2022objaverse} and is two orders of magnitude larger than ShapeNet.

The scale and diversity of assets in Objaverse-XL significantly expands the performance of state-of-the-art 3D models. The recently proposed Zero123~\cite{liu2023zero1to3} model for novel view synthesis, when pre-trained with Objaverse-XL, shows significantly better zero-shot generalization to challenging and complex modalities including photorealistic assets, cartoons, drawings and sketches. Similar improvements are also seen with PixelNerf which is trained to synthesize novel views given a small set of images. On each of these tasks, scaling pre-training data continues to show improvements from a thousand assets all the way up to 10 million, with few signs of slowing down, showing the promise and opportunities enabled with web scale data.

\section{Related Work}

\paragraph{Pre-training Datasets.}
Massive datasets have a prevalent role in modern, data-driven AI as they have produced powerful and general representations when paired with large-scale training. 
In computer vision, ImageNet~\cite{deng2009imagenet}, introduced nearly 14 years ago, has become the standard pre-training dataset of state-of-the-art visual models in object detection~\cite{ren2015faster,carion2020end}, instance segmentation~\cite{he2017mask,cheng2021mask2former} and more.
More recently, large image datasets, such as LAION-5B~\cite{schuhmann2022laion}, have powered exciting advances in generative AI, such as Stable Diffusion~\cite{rombach2022high}, and have given rise to new general-purpose vision and language representations with models like CLIP~\cite{radford2021learning} and Flamingo~\cite{alayrac2022flamingo}. This year, SAM~\cite{kirillov2023segment}

\begin{minipage}{\textwidth}
  \begin{minipage}[b]{0.7\textwidth}
    \centering
    \includegraphics[width=\linewidth]{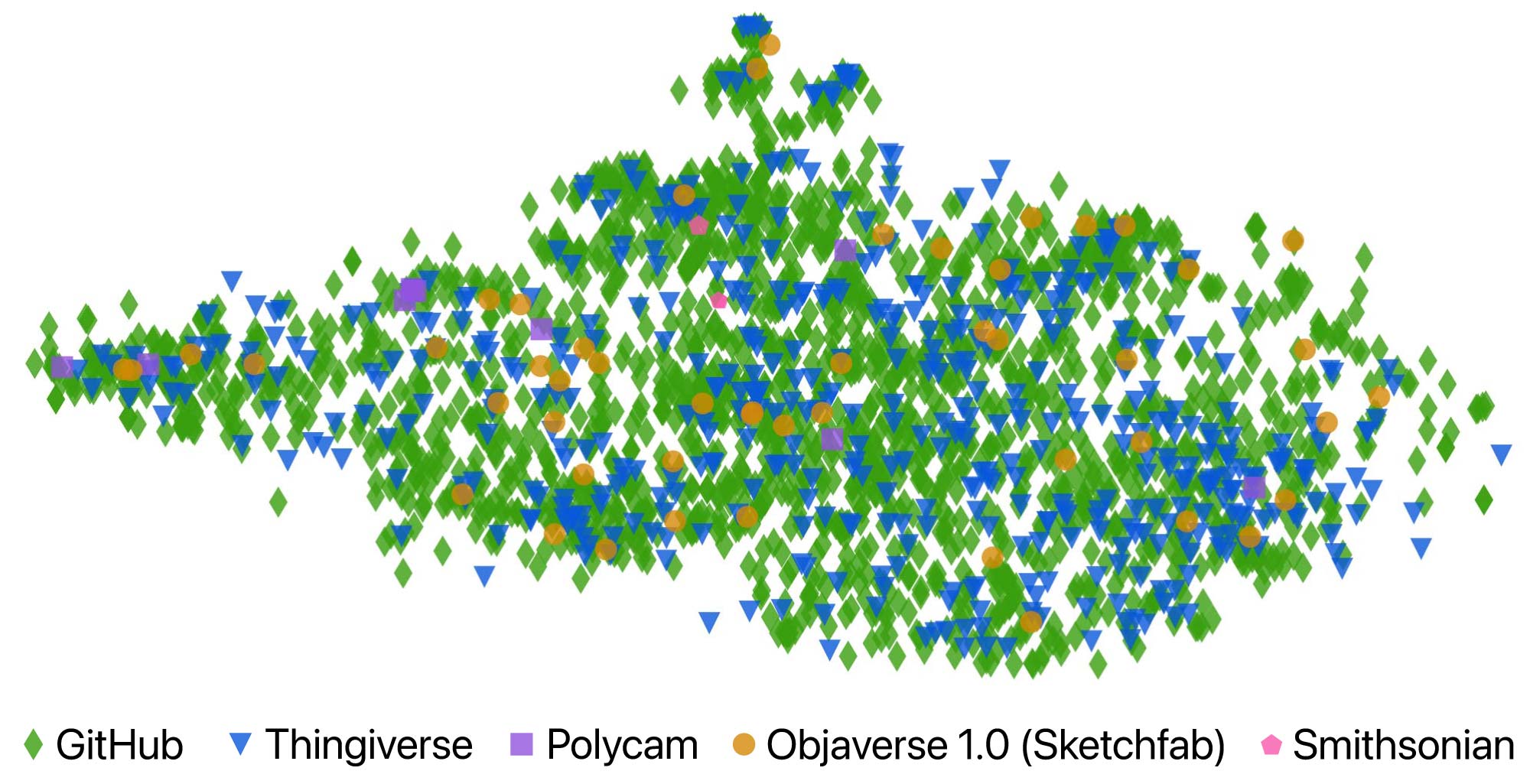}
    \captionof{figure}{t-SNE projection of CLIP L/14 embeddings on a subset of rendered objects. Compared to \datasetone (orange), \dataset more densely captures the distribution of 3D assets.}
    \label{fig:related}
  \end{minipage}
  \hfill
  \begin{minipage}[b]{0.275\textwidth}
    \centering
    \begin{adjustbox}{width=0.975\textwidth,center}
        \begin{tabular}{lr}
            \toprule
            Source & \# Objects\\
            \midrule
            IKEA~\cite{lim2013parsing} & 219\\
            GSO~\cite{downs2022google} & 1K\\
            EGAD~\cite{morrison2020egad} & 2K\\
            OmniObject3D~\cite{wu2023omniobject3d} & 6K\\
            PhotoShape~\cite{park2018photoshape} & 5K\\
            ABO~\cite{collins2022abo} & 8K\\
            Thingi10K~\cite{zhou2016thingi10k} & 10K\\
            3d-Future~\cite{fu20213d} & 10K\\
            ShapeNet~\cite{chang2015shapenet} & 51K\\\midrule
            Objaverse 1.0~\cite{deitke2022objaverse} & 800K\\[0.025in]
            \textbf{Objaverse-XL} & \textbf{10.2M}\\
            \bottomrule
        \end{tabular}
    \end{adjustbox}
    \vspace{1mm}
      \captionof{table}{Number of 3D models in common datasets. \dataset is over an order of magnitude larger than prior datasets.}
      \label{tab:related}
    \end{minipage}
\end{minipage}\vspace*{5mm}

introduced a dataset of one billion object masks used to train a model capable of segmenting any object from an image. 
In language understanding, datasets like Common Crawl~\cite{commoncrawl} have culminated in unprecedented capabilities of large language models such as GPT-4~\cite{openai2023gpt}, which in turn power mainstream applications like ChatGPT.
The impact of large datasets is undeniable. 
However, current efforts to collect massive datasets focus on image and language modalities. 
In this work we introduce and release publically a massive dataset of 3D objects, called \dataset. 
Given the promise of large datasets for 2D vision and language, we believe \dataset will accelerate research in large-scale training for 3D understanding.

\paragraph{3D Datasets.}
Existing 3D datasets have been instrumental in yielding significant findings in 3D over the years. 
ShapeNet~\cite{chang2015shapenet} has served as the testbed for modeling, representing and predicting 3D shapes in the era of deep learning. 
ShapeNet provides a collection of 3D shapes, in the form of textured CAD models labeled with semantic categories from WordNet~\cite{miller1995wordnet}. 
In theory, it contains 3M CAD models with textures. In practice, a small subset of 51K models is used after filtering by mesh and texture quality. 
Notwithstanding its impact, ShapeNet objects are of low resolution and textures are often overly simplistic.
Other datasets such as ABO~\cite{collins2022abo}, GSO~\cite{downs2022google}, and OmniObjects3D~\cite{wu2023omniobject3d} improve on the texture quality of their CAD models but are significantly smaller in size with the largest constituting 15K CAD models.
Recently, \datasetone~\cite{deitke2022objaverse} introduced a 3D dataset of 800K 3D models with high quality and diverse textures, geometry and object types, making it 15$\times$ larger than prior 3D datasets. 
While impressive and a step toward a large-scale 3D dataset, \datasetone remains several magnitudes smaller than dominant datasets in vision and language. 
As seen in Figure~\ref{fig:related} and Table~\ref{tab:related}, \dataset extends \datasetone to an even larger 3D dataset of $10.2$M unique objects from a diverse set of sources, object shapes, and categories. 
We discuss \dataset and its properties in Section~\ref{sec:obaverse}.

\begin{figure}
    \centering
    \includegraphics[width=1\textwidth]{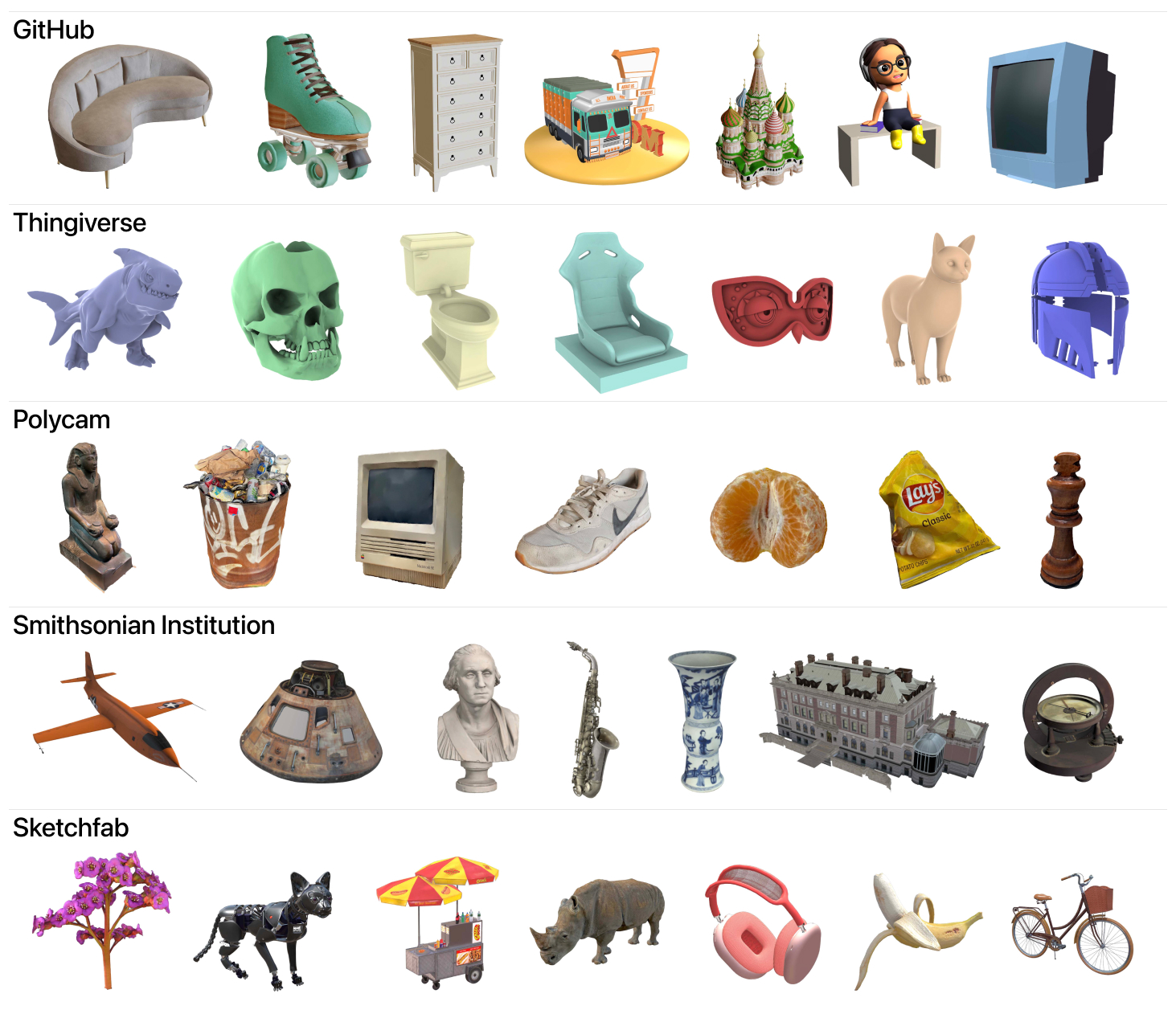}
    \caption{\textbf{Examples of 3D objects from various sources of \dataset} spanning GitHub, Thingiverse, Polycam, the Smithsonian Institution, and Sketchfab. Objects from Thingiverse do not include color information, so each object's primary color is randomized during rendering.}
\end{figure}

\paragraph{3D Applications.}
The potential of a massive 3D dataset like \dataset promises exciting novel applications in computer vision, graphics, augmented reality and generative AI. 
Reconstructing 3D objects from images is a longstanding problem in computer vision and graphics. 
Here, several methods explore novel representations~\cite{choy20163d,wang2018pixel2mesh,mescheder2019occupancy,mildenhall2020nerf}, network architectures~\cite{gkioxari2019mesh,yu2021pixelnerf} and differentiable rendering techniques~\cite{kato2018neural,chen2019learning,ravi2020accelerating,liu2023humans,liu2022shadows} to predict the 3D shapes and textures of objects from images with or without 3D supervision. 
All of the aforementioned projects experiment on the small scale ShapeNet. 
The significantly larger \dataset could pave the way to new levels of performance, and increase generalization to new domains in a zero-shot fashion. 
Over the past year, generative AI has made its foray into 3D. MCC~\cite{wu2023multiview} learns a generalizable representation with self-supervised learning for 3D reconstruction.
DreamFusion~\cite{poole2022dreamfusion} and later on Magic3D~\cite{lin2023magic3d} demonstrated that 3D shapes could be generated from language prompts with the help of text-to-image models. 
Point-E~\cite{nichol2022point} and Shape-E~\cite{jun2023shap} also train for text-to-3D with the help of 3D models from an undisclosed source. 
Recently, Zero123~\cite{liu2023zero1to3} introduced an image-conditioned diffusion model which generates novel object views and is trained on \datasetone. 
Stable Dreamfusion~\cite{stable-dreamfusion} replaces the text-to-image model in DreamFusion with the 3D-informed Zero123 and shows improved 3D generations. 
Recent findings in AI and scaling laws~\cite{kaplan2020scaling,hoffmann2022training} suggest that both generative and predictive models benefit from larger models and larger pre-training datasets. 
For 3D, \dataset is by far the largest 3D dataset to date and has the potential to facilitate large-scale training for new applications in 3D.

\section{Objaverse-XL}
\label{sec:obaverse}

\dataset is a web scale 3D object dataset composed of a highly diverse set of 3D data sources on the internet. In this section, we discuss the sources, metadata of the objects, and provide an analysis of the objects.

\subsection{Composition}

\dataset is composed of 3D objects coming from several sources, including GitHub, Thingiverse, Sketchfab, Polycam, and the Smithsonian Institution. We detail each source below.

\begin{figure}
    \begin{subfigure}{0.43\textwidth}
        \centering
        \includegraphics[width=0.9\textwidth]{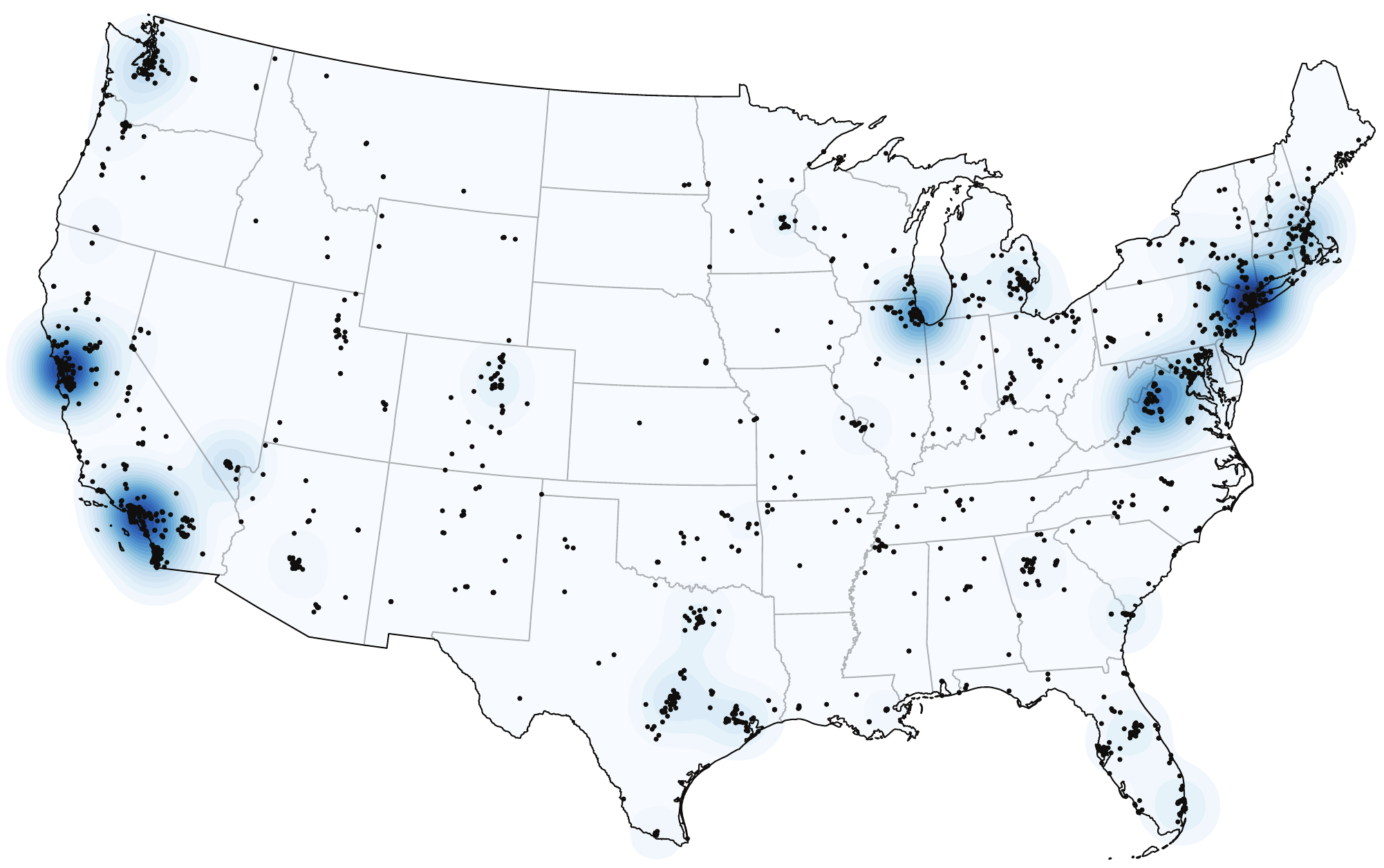}
        \caption{Object location density in the United States} %
        \label{fig:usdensity} %
    \end{subfigure}
    \hfill
    \begin{subfigure}{0.575\textwidth}
        \centering
        \includegraphics[width=\textwidth]{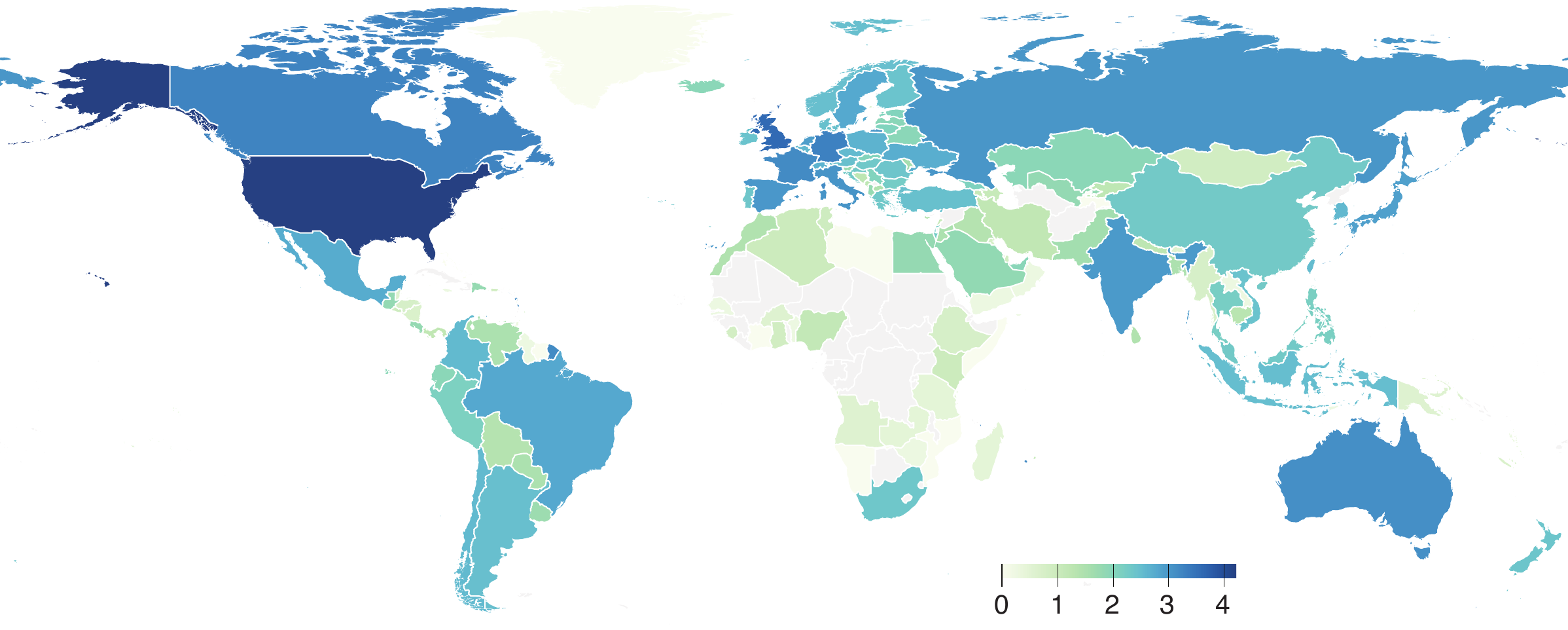}
        \caption{Choropleth map of objects per country (log scale)} %
        \label{fig:geotag5} %
    \end{subfigure}
    \\[0.10in]
    \begin{subfigure}{0.43\textwidth}
        \centering
        \includegraphics[width=\textwidth]{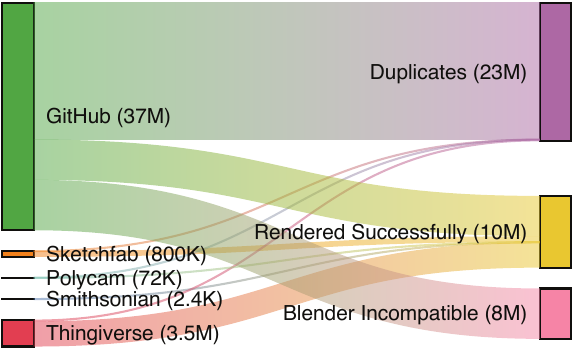}
        \caption{Sankey diagram of Objaverse-XL sources} %
        \label{fig:geotag5} %
    \end{subfigure}
    \hfill
    \begin{subfigure}{0.575\textwidth}
        \centering
        \includegraphics[width=0.95\textwidth]{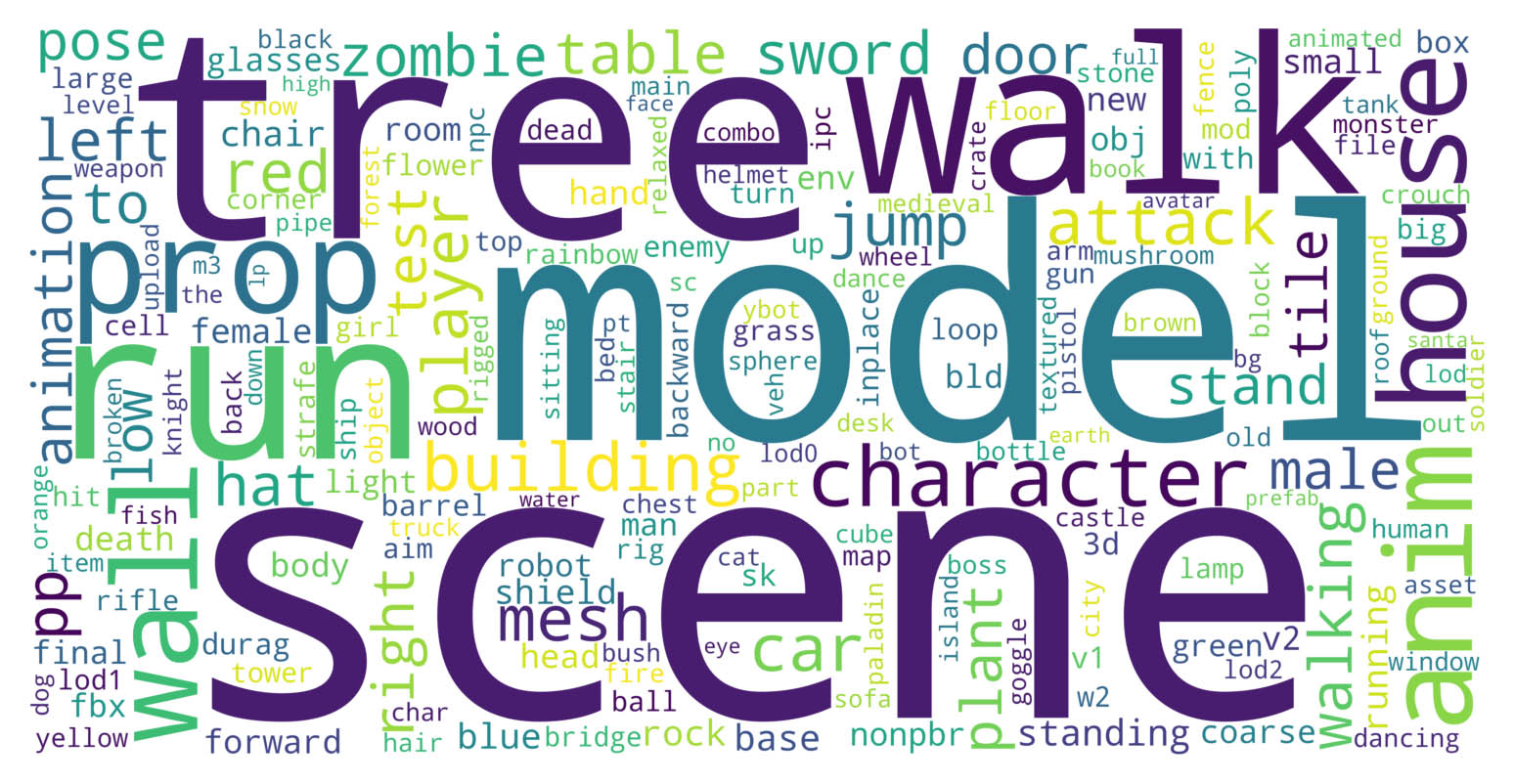}
        \caption{Word cloud of file names for GitHub} %
    \end{subfigure}
    \\[0.10in]
    \begin{subfigure}{\textwidth}
        \includegraphics[width=0.245\textwidth]{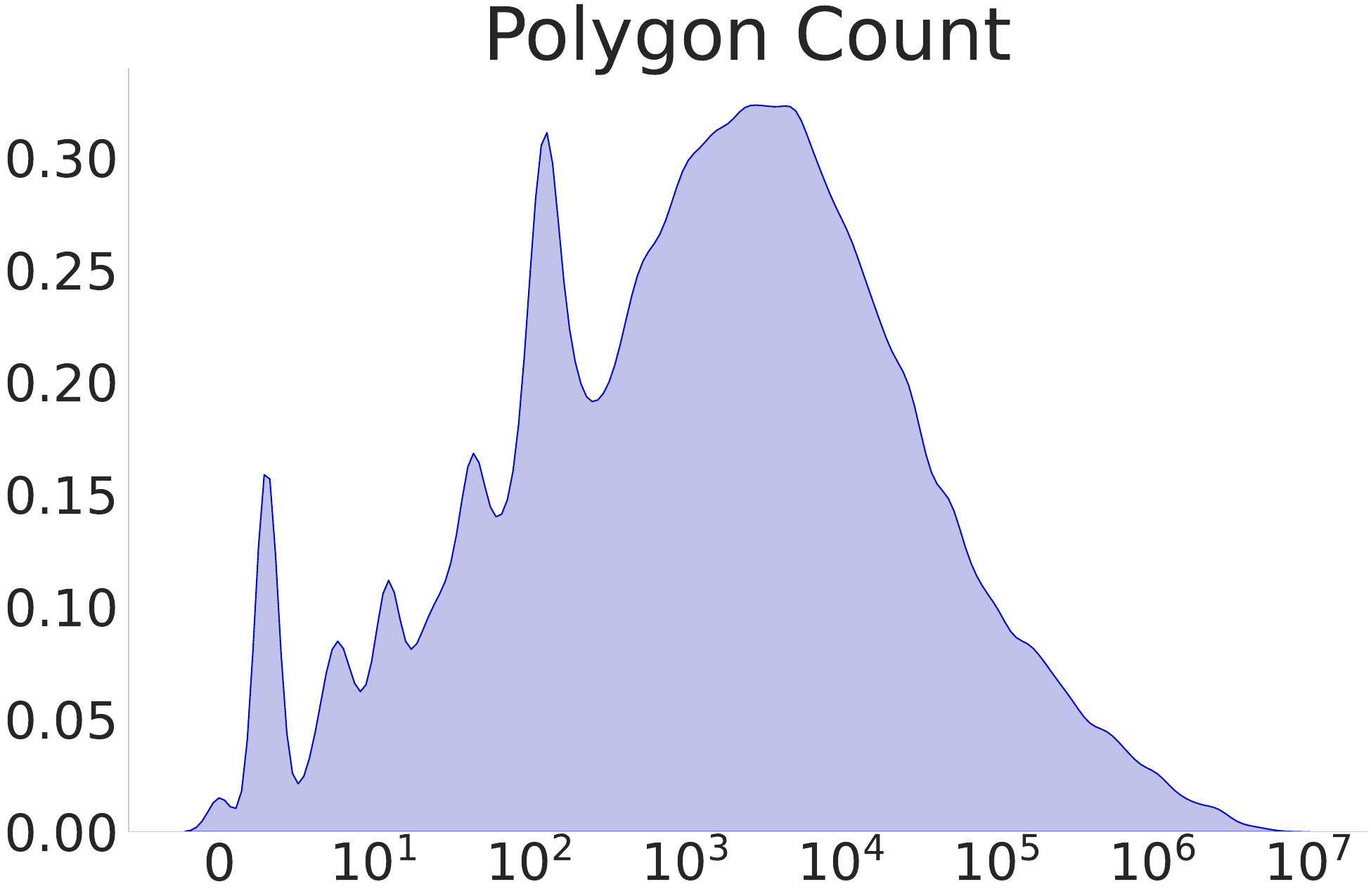}%
        \hfill%
        \includegraphics[width=0.245\textwidth]{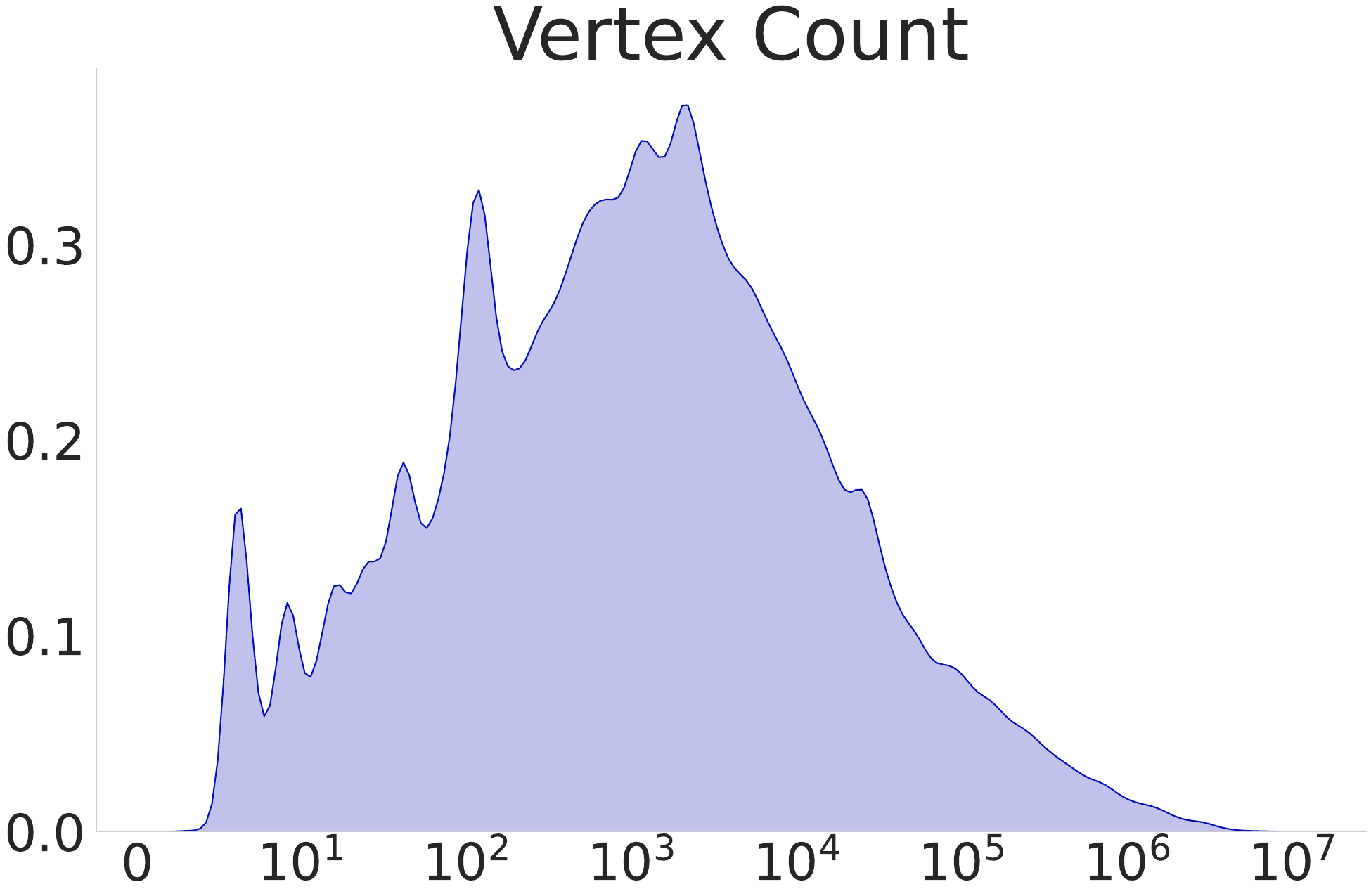}%
        \hfill%
        \includegraphics[width=0.245\textwidth]{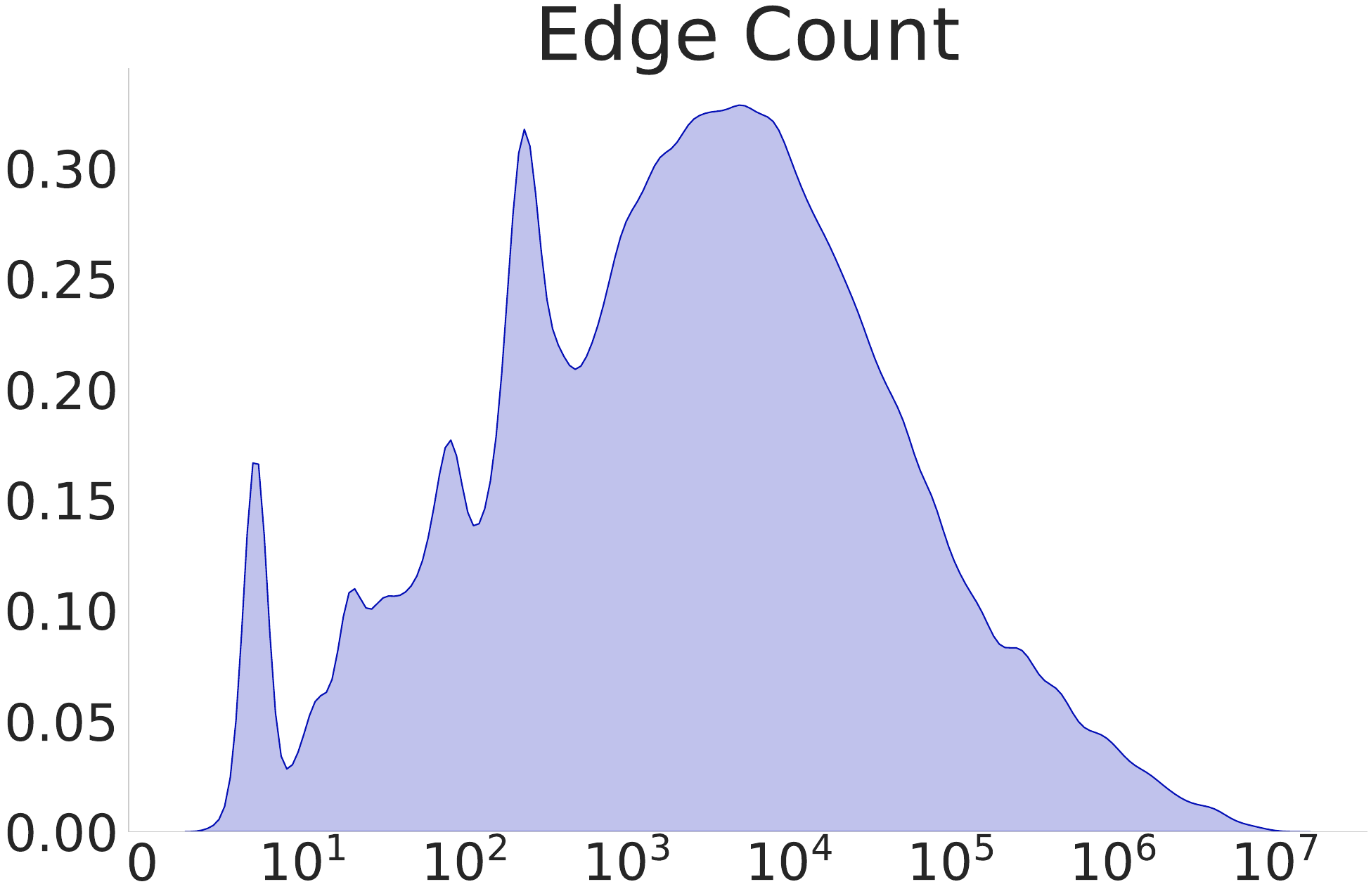}%
        \hfill%
        \includegraphics[width=0.245\textwidth]{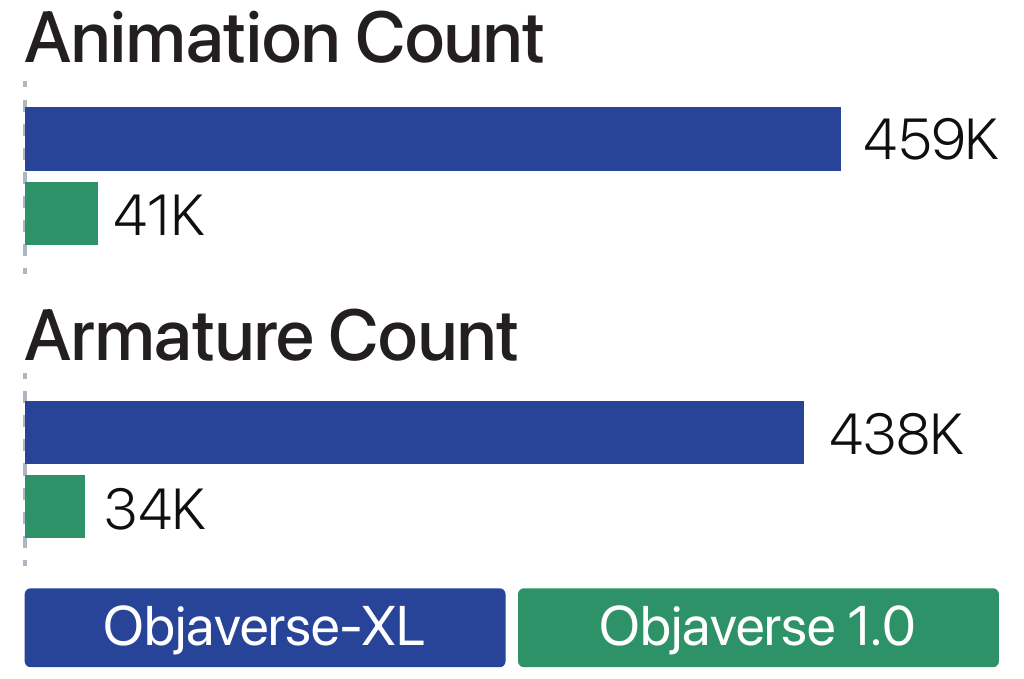}
        \caption{Statistics of the metadata extracted from the 3D objects.}
        \label{fig:statMeta}
    \end{subfigure}
    \caption{\textbf{Analysis of metadata from Objaverse-XL.} Locations of geotagged objects in (a) the United States and (b) around the world. (c) Various sources and their contribution to \dataset. (d) Frequency of filenames of GitHub objects. (e) Further statistics of collected 3D objects.}
    \label{fig:stats} %
\end{figure}

\textbf{GitHub} is a popular online platform for hosting code. We index 37M public files that contain common 3D object extensions; in particular, \texttt{.obj}, \texttt{.glb}, \texttt{.gltf}, \texttt{.usdz}, \texttt{.usd}, \texttt{.usda}, \texttt{.fbx}, \texttt{.stl}, \texttt{.dae}, \texttt{.ply}, \texttt{.abc}, and \texttt{.blend}. These extensions were chosen as they are best supported in Blender, which we use to render 2D images of the 3D objects. We only index objects that come from ``base'' GitHub repositories (\textit{i.e.} non-forked repos, excluding forks that had more stars than the original repo). In total, the files come from over 500K repositories.

Across all of \dataset, objects are deduplicated by file content hash, which removes approximately 23 million files. Among the remaining files, we were able to import and successfully render 5.5 million of those files. Files that were not successfully rendered were either caused by import compatibility issues (\textit{i.e.} FBX ASCII files are not natively importable to Blender), no meshes are in the files, or the file is not a valid 3D file (\textit{e.g.} an \texttt{.obj} file may be a C compiler file instead of a Wavefront Object file). Moving forward, we expect a solution for converting 3D file formats into a consolidated representation may yield several million more unique 3D objects.

\textbf{Thingiverse} is a platform for sharing objects most commonly used for 3D printing. We index and download around 3.5 million objects from the platform, which are predominantly released under Creative Commons licenses. The vast majority of the objects are STL files, which are often watertight meshes that are untextured, and serve as useful data for learning a shape prior. During rendering, we randomize the colors to broaden the distribution of the images.

\textbf{Sketchfab} is an online platform where users can publish and share 3D models, encompassing a broad variety of categories. The data sourced from Sketchfab for our project is specifically from Objaverse 1.0, a dataset of 800K objects consisting of Creative Commons-licensed 3D models. Each model is distributed as a standardized GLB file. The 3D models are freely usable and modifiable, covering an array of object types, from real-world 3D scans to intricate designs created in 3D software.

\textbf{Polycam} is a 3D scanning mobile application designed to facilitate the acquisition and sharing of 3D data. One of its salient features is the \textit{explore} functionality, which enables members of the user community to contribute their 3D scans to a publicly accessible database. In the context of our dataset, we focus specifically on the subset of objects within the explore page that are designated as savable. These savable objects are governed by a Creative Commons Attribution 4.0 International License (CC-BY 4.0). We indexed 72K objects that were marked as savable and licensed under a CC-BY 4.0 license. Following deduplication, we obtain 71K unique objects.

\textbf{Smithsonian 3D Digitization} is a project by the Smithsonian Institution dedicated to digitizing their vast collection of historical and cultural artifacts. The project has provided us with a set of 2.4K models, all licensed under a CC0 license, which signifies that these works are fully in the public domain and free for use without any restrictions. The objects in this collection are primarily scans of real-world artifacts. Each model is distributed in a standardized compressed GLB format.

\subsection{Metadata}
\label{sec:metadata}

Each object comes with metadata from its source, and we also extract metadata from it in Blender and from its CLIP ViT-L/14 features. We describe the metadata acquisition process below.

\paragraph{Source Metadata.} From the source, we often get metadata such as its popularity, license, and some textual description. For example, on GitHub, the popularity is represented by the stars of the object's repository and the file name serves as the object's textual pair.

\paragraph{Blender Metadata.} For each object that we render, we obtain the following metadata for it: \texttt{sha256}, \texttt{file-size}, \texttt{polygon-count}, \texttt{vertex-count}, \texttt{edge-count}, \texttt{material-count}, \texttt{texture-count}, \texttt{object-count}, \texttt{animation-count},
\texttt{linked-files},
\texttt{scene-dimensions}, and
\texttt{missing-textures}.
During rendering, for objects that have a missing texture, we randomize the color of that texture. Figure~\ref{fig:stats} shows some charts extracted from the metadata, including density plots over the number of polygons, vertex counts, and edge counts.%

\paragraph{Animated Objects.}
From the Blender metadata, we find that the number of animated objects and those with armature (a digital skeleton used to animate 3D models) significantly increases from Objaverse 1.0 to Objaverse-XL. Figure~\ref{fig:statMeta} (right) shows a bar chart of the increase, specifically from 41K to 459K animated objects and from 34K to 438K objects with armature.

\paragraph{Model Metadata.}
For each object, we extract its CLIP ViT-L/14~\cite{radford2021learning} image embedding by averaging the CLIP embedding from 12 different renders of the object at random camera positions inside of a hollow sphere. We use the CLIP embeddings to predict different metadata properties, including aesthetic scores, NSFW predictions, face detection, and for detecting holes in the photogrammetry renderings. Section~\ref{sec:analysis} provides more details on the analysis.

\begin{figure*}[t!]
    \centering
    \begin{adjustbox}{center}
        \includegraphics[width=1.225\textwidth]{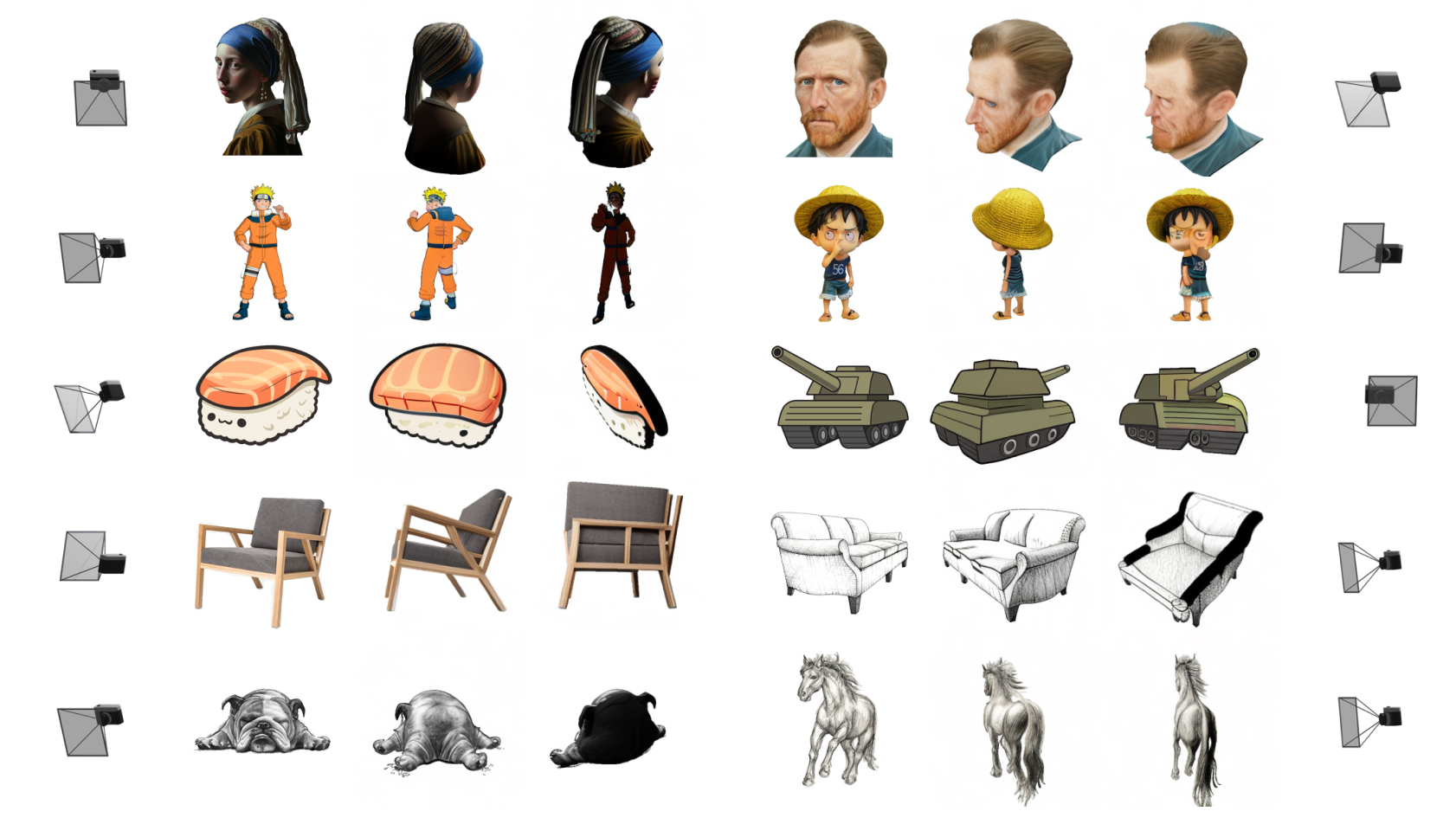}
    \end{adjustbox}\\[-0.10in]
    \begin{adjustbox}{center}
        \includegraphics[width=1.225\textwidth]{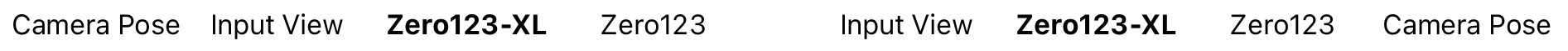}
    \end{adjustbox}\\[0.10in]
    \caption{\textbf{Novel view synthesis on in-the-wild images.} Comparison between Zero123-XL trained on Objaverse-XL and Zero123 trained on Objaverse. Starting from the input view, the task is to generate an image of the object under a specific camera pose transformation. The camera poses are shown beside each example. Significant improvement can be found by training with more data, especially for categories including people (\textbf{1\textsuperscript{st} row)}, anime (\textbf{2\textsuperscript{nd} row)}, cartoon (\textbf{3\textsuperscript{rd} row)}, furniture (\textbf{4\textsuperscript{th} row)}, and sketches (\textbf{5\textsuperscript{th} row)}. Additionally, viewpoint control is significantly improved (see \textbf{2\textsuperscript{nd} row)}.
    }
    \label{fig:zero123}
\end{figure*}

\subsection{Analysis}
\label{sec:analysis}

\paragraph{NSFW annotations.} Most data sources used for the creation of \dataset already have either a strict NSFW policy or strong self-filtering. However, owing to the web scale of \dataset we performed NSFW filtering using the rendered images of the objects. Each 3D object is rendered in 12 random views and each rendered image is passed through an NSFW classifier trained on the NSFW dataset introduced in LAION-5B~\citep{schuhmann2022laion} by~\citet{gadre2023datacomp} using the CLIP ViT-L/14~\citep{radford2021learning} features. After careful analysis and manual inspection, we marked a rendered image as NSFW if it has an NSFW score above $0.9$ and a 3D object is marked as NSFW if at least $3$ rendered images are deemed to be NSFW. Overall, only $815$ objects out of the 10M are filtered out as NSFW objects. Note that the high threshold and multi-view consistency are needed due to the distribution shift between LAION-5B and \dataset along with NSFW classification of certain viewpoint renders of harmless 3D objects.

\paragraph{Face detection.} We analyze the presence of faces in \dataset using a detector trained by~\citet{gadre2023datacomp}. Like NSFW filtering, we count the objects where at least $3$ images contain a detected face. Out of 10M assets, we estimate $266$K objects include faces. However, unlike most web-scale datasets, the faces present in \dataset often come from the scans of dolls, historical sculptures, and anthropomorphic animations. Hence, there are less privacy concerns with most of these objects.

\paragraph{Photogrammetry hole detection.} When scanning 3D objects, if the back or bottom of the object is not scanned, rendering from various viewpoints may contain holes, leading to a ``bad'' render image. For example, a non-trivial number of Polycam 3D objects lack the information from the ``back side''. In most cases, images that are rendered from back-side viewpoints are noisy, low-fidelity, or contain holes. To analyze ``bad rendering'' at scale, we manually annotated 1.2K Polycam renders as ``good'' (label $1$) or ``bad'' (label $0$). We trained a ``bad render'' classifier (2-layer MLP) on top of the CLIP ViT-L/14 features of the rendered images; this classifier achieves a cross-validation accuracy of over $90\%$ with a ``render score'' threshold of $0.5$. Overall, out of $71$K Polycam objects with 12 renders each, we found that $38.20\%$ renders are ``bad'', with  $58$K objects having at least 2 bad renders.

\section{Experiments}
\label{sec:experiments}

\subsection{Novel View Synthesis with Zero123-XL}

\begin{figure*}
    \centering
    \includegraphics[width=0.95\textwidth]{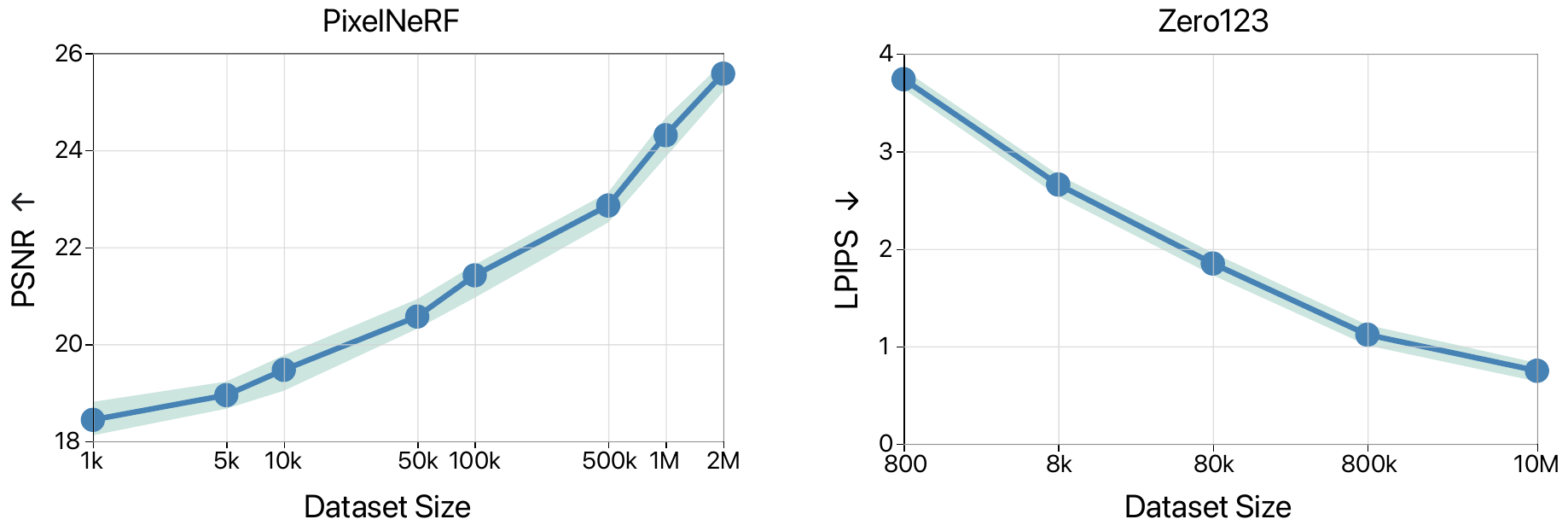}
    \captionof{figure}{\textbf{Novel view synthesis at scale.} \textbf{Left}: PixelNeRF~\cite{yu2021pixelnerf} trained on varying scales of data and evaluated on a held-out subset of Objavserse-XL. \textbf{Right}: Zero123~\cite{liu2023zero1to3} trained on varying scales of data and evaluated on a zero-shot dataset.  Note that the 800K datapoint is  Zero123 and the 10M datapoint is Zero123-XL. The synthesis quality consistently improves with scale. LPIPS is scaled up 10 times for visualization.}
    \label{fig:scale}
    \vspace*{-5mm}
\end{figure*}

\begin{table}[b!]
\centering
\vspace*{-4mm}
\begin{tabular}{@{}ccccc@{}}
\toprule
Zero123-XL                 & PSNR ($\uparrow$)    & SSIM ($\uparrow$) & LPIPS ($\downarrow$) & FID ($\downarrow$) \\ \midrule
{Base} & 18.225             & 0.877           & 0.088              & 0.070                        \\
{w/ Alignment Finetuning}               & \textbf{19.876}             & \textbf{0.888 }          & \textbf{0.075}              & \textbf{0.056}                        \\
 \bottomrule
\end{tabular}
\vspace{0.05in}
\caption{\textbf{Effect of high-quality data finetuning on Zero123-XL.} When evaluated zero-shot on Google Scanned Objects~\cite{downs2022google}, a model finetuned on a high-quality alignment subset of \dataset significantly outperforms the base model trained only on \dataset.
}

\label{tab:alignment}
\end{table}

Generating 3D assets conditioned on in-the-wild 2D images has remained a challenging problem in computer vision. A crucial lesson learned from large language models is that pretraining on simple and easily scalable tasks, such as next word prediction, leads to consistently improved performance and the emergence of zero-shot abilities. An analogous approach in 3D vision is to predict a novel view of an object from an input view. Zero123~\cite{liu2023zero1to3} recently proposed a view-conditioned diffusion model to perform this task, where the weights of the diffusion model are initialized from Stable Diffusion to leverage its powerful zero-shot image generation abilities. Zero123 used objects in Objaverse 1.0 to render input and novel view pairs as the training dataset.
We use this framework to create \textit{Zero123-XL}, which is the same approach except trained on the much larger Objaverse-XL instead. As shown in~\cite{liu2023zero1to3}, the pretrained view-conditioned diffusion model can also be plugged into a score distillation framework such as DreamFusion~\cite{poole2022dreamfusion} or SJC~\cite{wang2023score} to obtain a 3D assets.

\paragraph{Zero-shot Generalization.} We found that training Zero123 on Objaverse-XL achieves significantly better zero-shot generalization performance than using Objaverse 1.0. Figure~\ref{fig:zero123} shows examples from categories of data commonly known to be challenging for baseline systems, including people, cartoons, paintings, and sketches. For example, in both of the examples shown in 2nd and 3rd rows of the first column, Zero123 interprets the input image as a 2D plane and performs a simple transformation similar to a homography transformation. In comparison, Zero123-XL is able to generate novel views that are more consistent with the input view. Additionally, Zero123-XL is able to generate novel views from sketches of objects while keeping the original style as well as object geometric details. These examples show the effectiveness of dataset scaling for zero-shot generalization in 3D.

\paragraph{Improvement with Scale.} We further quantitatively evaluate the novel view synthesis performance on Google Scanned Objects dataset~\cite{downs2022google}. As shown in Figure~\ref{fig:scale}, the rvisual similarity score~\cite{zhang2018unreasonable} between the predicted novel view and the ground truth view continues to improve as the dataset size increases.

\paragraph{Alignment Finetuning.}
InstructGPT~\cite{ouyang2022training} shows that large-scale pretraining does not directly lead to a model aligned with human preferences. More recently, LIMA~\cite{zhou2023lima} shows that finetuning a pretrained model on a curated subset with high-quality data can achieve impressive alignment results. We adopted a similar approach here by selecting a high-quality subset of Objaverse-XL that contains 1.3 million objects. Selection is done by defining proxy estimation of human preference based on heuristics including vertex count, face count, popularity on the source website, and source of data, among other metrics. After pretraining the base model on the entire Objaverse-XL, we finetune Zero123-XL on the alignment subset with a reduced learning rate and performed an ablation study to evaluate the effect of alignment finetuning. Table~\ref{tab:alignment} shows that alignment finetuning leads to significant improvement in zero-shot generalization performance. Please refer to Appendix~\ref{sec:appendixImpl} for more implementation details regarding our model and experiments.

\subsection{Novel View Synthesis with PixelNeRF}
Synthesizing objects and scenes from novel views is a long-standing challenge. Notably, neural radiance fields \cite{mildenhall2020nerf} have shown impressive capabilities in rendering specific scenes from novel views. However, these methods require dozens of views of an individual scene, and can only synthesize views from the particular scene they were trained for.  More recent methods \cite{deng2022depth, jain2021putting, wang2021ibrnet, yu2021pixelnerf} have been proposed for constructing NeRF models that generalize across scenes with few input images. Due to the challenging nature of obtaining the necessary camera parameters for training, such methods have traditionally been trained on small scale data sets. With the Objaverse-XL data, we train a PixelNeRF model on over two million objects, magnitudes of more data than has previously been used. We find that PixelNeRF generalizes to novel scenes and objects significantly better and performance improves consistently with scale (Figure \ref{fig:scale} and Table \ref{tab:fine-tune}).

\paragraph{Improvement with Scale.} We train PixelNeRF models conditioned on a single input image at varying scales of data (Figure \ref{fig:scale}) and evaluate on a held out set of Objaverse-XL objects. We find that novel view synthesis quality consistently improves with more objects even at the scale of 2 million objects and 24 million rendered images.

\setlength{\textfloatsep}{5pt}
\begin{wrapfigure}{r}{0.5\columnwidth}\vspace{-4mm}
\centering
\begin{tabular}{@{}ccc@{}}
\toprule
                   PixelNeRF                            & {DTU~\citep{aanaes2016large}} & ShapeNet~\citep{chang2015shapenet} \\ \midrule
{Base}                 & 15.32                   & 22.71                        \\
{w/ \dataset} & \textbf{17.53 $\pm .37$}                   & \textbf{24.22 $\pm .55$ }                       \\ \bottomrule
\end{tabular}

\captionof{table}{\small\textbf{Comparison (PSNR ($\uparrow$)) of PixelNeRF trained from scratch vs. fine-tuned from Objaverse-XL.} Performance significantly improves from pretraining on the large-scale corpus.}
\label{tab:fine-tune}
  \vspace{-4mm}
\end{wrapfigure}
\paragraph{Generalization to Downstream Datasets.} 
Similar to pretraining in 2D vision and language, we observe that pretraining on Objaverse-XL with PixelNeRF improves performance when fine-tuning to other datasets such as DTU \cite{aanaes2016large} and ShapeNet \cite{chang2015shapenet} (Table \ref{tab:fine-tune}). We pretrain and fine-tune the model conditioned on a single input view and report the peak signal-to-noise ratio (PSNR).

\vspace*{-2mm}
\section{Limitations and Conclusion}
\label{sec:conclusion}

\paragraph{Limitations.}
While \dataset is more than an order of magnitude larger than its predecessor, \datasetone, it is still orders of magnitude smaller than modern billion-scale image-text datasets. Future work may consider how to continue scaling 3D datasets and make 3D content easier to capture and create. Additionally, it may not be the case that all samples in \dataset are necessary to train high performance models. Future work may also consider how to choose datapoints to train on. Finally, while we focus on generative tasks, future work may consider how \dataset can benefit discriminative tasks such as 3D segmentation and detection.

\vspace*{-2mm}
\paragraph{Conclusion.}
We introduce \dataset, which is comprised of 10.2M 3D assets.
In addition to documenting \dataset's unprecedented scale and sample diversity, we demonstrate the potential of \dataset for downstream applications.
On the task of zero-shot novel view synthesis, we establish empirically promising trends of scaling dataset size, while keeping the model architecture constant.
We hope \dataset will provide a foundation for future work in 3D.
\clearpage

\section*{Acknowledgements}

We would like to thank Stability AI for compute used to train the experiments and LAION for their support. We would also like to thank Luca Weihs, Mitchell Wortsman, Romain Beaumont, and Vaishaal Shankar, Rose Hendrix, Adam Letts, Sami Kama, Andreas Blattmann, Kunal Pratap Singh, and Kuo-Hao Zeng for their helpful guidance and conversations with the project. Finally, we would like to thank the teams behind several open-source packages used throughout this project, including Blender~\cite{blender}, PyTorch~\cite{paszke2019pytorch}, PyTorch Lightning~\cite{Falcon_PyTorch_Lightning_2019}, D3~\cite{bostock2011d3}, Matplotlib~\cite{Hunter:2007}, NumPy~\cite{harris2020array}, Pandas~\cite{reback2020pandas}, Wandb~\cite{wandb}, and Seaborn~\cite{Waskom2021}. Following the NeurIPS guidelines, we would also like to acknowledge the use of LLMs for helping revise some text and general coding assistance. Finally, we would also like to thank and acknowledge the content creators who contributed to the dataset.

\bibliography{egbib}

\begin{thebibliography}{67}
\providecommand{\natexlab}[1]{#1}
\providecommand{\url}[1]{\texttt{#1}}
\expandafter\ifx\csname urlstyle\endcsname\relax
  \providecommand{\doi}[1]{doi: #1}\else
  \providecommand{\doi}{doi: \begingroup \urlstyle{rm}\Url}\fi

\bibitem[com()]{commoncrawl}
URL \url{https://commoncrawl.org/the-data/}.

\bibitem[Aan{\ae}s et~al.(2016)Aan{\ae}s, Jensen, Vogiatzis, Tola, and
  Dahl]{aanaes2016large}
H.~Aan{\ae}s, R.~R. Jensen, G.~Vogiatzis, E.~Tola, and A.~B. Dahl.
\newblock Large-scale data for multiple-view stereopsis.
\newblock \emph{International Journal of Computer Vision}, pages 1--16, 2016.

\bibitem[Alayrac et~al.(2022)Alayrac, Donahue, Luc, Miech, Barr, Hasson, Lenc,
  Mensch, Millican, Reynolds, et~al.]{alayrac2022flamingo}
J.-B. Alayrac, J.~Donahue, P.~Luc, A.~Miech, I.~Barr, Y.~Hasson, K.~Lenc,
  A.~Mensch, K.~Millican, M.~Reynolds, et~al.
\newblock Flamingo: a visual language model for few-shot learning.
\newblock \emph{Advances in Neural Information Processing Systems},
  35:\penalty0 23716--23736, 2022.

\bibitem[Biewald(2020)]{wandb}
L.~Biewald.
\newblock Experiment tracking with weights and biases, 2020.
\newblock URL \url{https://www.wandb.com/}.
\newblock Software available from wandb.com.

\bibitem[{Blender Online Community}(2023)]{blender}
{Blender Online Community}.
\newblock Blender - a 3d modelling and rendering package.
\newblock \url{https://www.blender.org}, 2023.

\bibitem[Bostock et~al.(2011)Bostock, Ogievetsky, and Heer]{bostock2011d3}
M.~Bostock, V.~Ogievetsky, and J.~Heer.
\newblock D3: Data-driven documents.
\newblock \emph{IEEE Transactions on Visualization and Computer Graphics},
  2011.

\bibitem[Brown et~al.(2020)Brown, Mann, Ryder, Subbiah, Kaplan, Dhariwal,
  Neelakantan, Shyam, Sastry, Askell, et~al.]{brown2020language}
T.~Brown, B.~Mann, N.~Ryder, M.~Subbiah, J.~D. Kaplan, P.~Dhariwal,
  A.~Neelakantan, P.~Shyam, G.~Sastry, A.~Askell, et~al.
\newblock Language models are few-shot learners.
\newblock \emph{Advances in neural information processing systems},
  33:\penalty0 1877--1901, 2020.

\bibitem[Carion et~al.(2020)Carion, Massa, Synnaeve, Usunier, Kirillov, and
  Zagoruyko]{carion2020end}
N.~Carion, F.~Massa, G.~Synnaeve, N.~Usunier, A.~Kirillov, and S.~Zagoruyko.
\newblock End-to-end object detection with transformers.
\newblock In \emph{Computer Vision--ECCV 2020: 16th European Conference,
  Glasgow, UK, August 23--28, 2020, Proceedings, Part I 16}, pages 213--229.
  Springer, 2020.

\bibitem[Chang et~al.(2015)Chang, Funkhouser, Guibas, Hanrahan, Huang, Li,
  Savarese, Savva, Song, Su, et~al.]{chang2015shapenet}
A.~X. Chang, T.~Funkhouser, L.~Guibas, P.~Hanrahan, Q.~Huang, Z.~Li,
  S.~Savarese, M.~Savva, S.~Song, H.~Su, et~al.
\newblock Shapenet: An information-rich 3d model repository.
\newblock \emph{arXiv preprint arXiv:1512.03012}, 2015.

\bibitem[Chen et~al.(2019)Chen, Ling, Gao, Smith, Lehtinen, Jacobson, and
  Fidler]{chen2019learning}
W.~Chen, H.~Ling, J.~Gao, E.~Smith, J.~Lehtinen, A.~Jacobson, and S.~Fidler.
\newblock Learning to predict 3d objects with an interpolation-based
  differentiable renderer.
\newblock \emph{Advances in neural information processing systems}, 32, 2019.

\bibitem[Cheng et~al.(2022)Cheng, Misra, Schwing, Kirillov, and
  Girdhar]{cheng2021mask2former}
B.~Cheng, I.~Misra, A.~G. Schwing, A.~Kirillov, and R.~Girdhar.
\newblock Masked-attention mask transformer for universal image segmentation.
\newblock 2022.

\bibitem[Choy et~al.(2016)Choy, Xu, Gwak, Chen, and Savarese]{choy20163d}
C.~B. Choy, D.~Xu, J.~Gwak, K.~Chen, and S.~Savarese.
\newblock 3d-r2n2: A unified approach for single and multi-view 3d object
  reconstruction.
\newblock In \emph{Computer Vision--ECCV 2016: 14th European Conference,
  Amsterdam, The Netherlands, October 11-14, 2016, Proceedings, Part VIII 14},
  pages 628--644. Springer, 2016.

\bibitem[Collins et~al.(2022)Collins, Goel, Deng, Luthra, Xu, Gundogdu, Zhang,
  Vicente, Dideriksen, Arora, et~al.]{collins2022abo}
J.~Collins, S.~Goel, K.~Deng, A.~Luthra, L.~Xu, E.~Gundogdu, X.~Zhang, T.~F.~Y.
  Vicente, T.~Dideriksen, H.~Arora, et~al.
\newblock Abo: Dataset and benchmarks for real-world 3d object understanding.
\newblock In \emph{Proceedings of the IEEE/CVF Conference on Computer Vision
  and Pattern Recognition}, pages 21126--21136, 2022.

\bibitem[Deitke et~al.(2022)Deitke, Schwenk, Salvador, Weihs, Michel,
  VanderBilt, Schmidt, Ehsani, Kembhavi, and Farhadi]{deitke2022objaverse}
M.~Deitke, D.~Schwenk, J.~Salvador, L.~Weihs, O.~Michel, E.~VanderBilt,
  L.~Schmidt, K.~Ehsani, A.~Kembhavi, and A.~Farhadi.
\newblock Objaverse: A universe of annotated 3d objects.
\newblock \emph{arXiv preprint arXiv:2212.08051}, 2022.

\bibitem[Deng et~al.(2009)Deng, Dong, Socher, Li, Li, and
  Fei-Fei]{deng2009imagenet}
J.~Deng, W.~Dong, R.~Socher, L.-J. Li, K.~Li, and L.~Fei-Fei.
\newblock Imagenet: A large-scale hierarchical image database.
\newblock In \emph{2009 IEEE conference on computer vision and pattern
  recognition}, pages 248--255. Ieee, 2009.

\bibitem[Deng et~al.(2022)Deng, Liu, Zhu, and Ramanan]{deng2022depth}
K.~Deng, A.~Liu, J.-Y. Zhu, and D.~Ramanan.
\newblock Depth-supervised nerf: Fewer views and faster training for free.
\newblock In \emph{Proceedings of the IEEE/CVF Conference on Computer Vision
  and Pattern Recognition}, pages 12882--12891, 2022.

\bibitem[Downs et~al.(2022)Downs, Francis, Koenig, Kinman, Hickman, Reymann,
  McHugh, and Vanhoucke]{downs2022google}
L.~Downs, A.~Francis, N.~Koenig, B.~Kinman, R.~Hickman, K.~Reymann, T.~B.
  McHugh, and V.~Vanhoucke.
\newblock Google scanned objects: A high-quality dataset of 3d scanned
  household items.
\newblock In \emph{2022 International Conference on Robotics and Automation
  (ICRA)}, pages 2553--2560. IEEE, 2022.

\bibitem[Falcon and {The PyTorch Lightning
  team}(2019)]{Falcon_PyTorch_Lightning_2019}
W.~Falcon and {The PyTorch Lightning team}.
\newblock {PyTorch Lightning}, Mar. 2019.
\newblock URL \url{https://github.com/Lightning-AI/lightning}.

\bibitem[Fu et~al.(2021)Fu, Jia, Gao, Gong, Zhao, Maybank, and Tao]{fu20213d}
H.~Fu, R.~Jia, L.~Gao, M.~Gong, B.~Zhao, S.~Maybank, and D.~Tao.
\newblock 3d-future: 3d furniture shape with texture.
\newblock \emph{International Journal of Computer Vision}, 129:\penalty0
  3313--3337, 2021.

\bibitem[Gadre et~al.(2023)Gadre, Ilharco, Fang, Hayase, Smyrnis, Nguyen,
  Marten, Wortsman, Ghosh, Zhang, et~al.]{gadre2023datacomp}
S.~Y. Gadre, G.~Ilharco, A.~Fang, J.~Hayase, G.~Smyrnis, T.~Nguyen, R.~Marten,
  M.~Wortsman, D.~Ghosh, J.~Zhang, et~al.
\newblock Datacomp: In search of the next generation of multimodal datasets.
\newblock \emph{arXiv preprint arXiv:2304.14108}, 2023.

\bibitem[Gebru et~al.(2021)Gebru, Morgenstern, Vecchione, Vaughan, Wallach,
  Iii, and Crawford]{gebru2021datasheets}
T.~Gebru, J.~Morgenstern, B.~Vecchione, J.~W. Vaughan, H.~Wallach, H.~D. Iii,
  and K.~Crawford.
\newblock Datasheets for datasets.
\newblock \emph{Communications of the ACM}, 64\penalty0 (12):\penalty0 86--92,
  2021.

\bibitem[Gkioxari et~al.(2019)Gkioxari, Malik, and Johnson]{gkioxari2019mesh}
G.~Gkioxari, J.~Malik, and J.~Johnson.
\newblock Mesh r-cnn.
\newblock In \emph{Proceedings of the IEEE/CVF International Conference on
  Computer Vision}, pages 9785--9795, 2019.

\bibitem[Harris et~al.(2020)Harris, Millman, van~der Walt, Gommers, Virtanen,
  Cournapeau, Wieser, Taylor, Berg, Smith, Kern, Picus, Hoyer, van Kerkwijk,
  Brett, Haldane, del R{\'{i}}o, Wiebe, Peterson, G{\'{e}}rard-Marchant,
  Sheppard, Reddy, Weckesser, Abbasi, Gohlke, and Oliphant]{harris2020array}
C.~R. Harris, K.~J. Millman, S.~J. van~der Walt, R.~Gommers, P.~Virtanen,
  D.~Cournapeau, E.~Wieser, J.~Taylor, S.~Berg, N.~J. Smith, R.~Kern, M.~Picus,
  S.~Hoyer, M.~H. van Kerkwijk, M.~Brett, A.~Haldane, J.~F. del R{\'{i}}o,
  M.~Wiebe, P.~Peterson, P.~G{\'{e}}rard-Marchant, K.~Sheppard, T.~Reddy,
  W.~Weckesser, H.~Abbasi, C.~Gohlke, and T.~E. Oliphant.
\newblock Array programming with {NumPy}.
\newblock \emph{Nature}, 585\penalty0 (7825):\penalty0 357--362, Sept. 2020.
\newblock \doi{10.1038/s41586-020-2649-2}.
\newblock URL \url{https://doi.org/10.1038/s41586-020-2649-2}.

\bibitem[He et~al.(2017)He, Gkioxari, Doll{\'a}r, and Girshick]{he2017mask}
K.~He, G.~Gkioxari, P.~Doll{\'a}r, and R.~Girshick.
\newblock Mask r-cnn.
\newblock In \emph{Proceedings of the IEEE international conference on computer
  vision}, pages 2961--2969, 2017.

\bibitem[Hoffmann et~al.(2022)Hoffmann, Borgeaud, Mensch, Buchatskaya, Cai,
  Rutherford, Casas, Hendricks, Welbl, Clark, et~al.]{hoffmann2022training}
J.~Hoffmann, S.~Borgeaud, A.~Mensch, E.~Buchatskaya, T.~Cai, E.~Rutherford,
  D.~d.~L. Casas, L.~A. Hendricks, J.~Welbl, A.~Clark, et~al.
\newblock Training compute-optimal large language models.
\newblock \emph{arXiv preprint arXiv:2203.15556}, 2022.

\bibitem[Hunter(2007)]{Hunter:2007}
J.~D. Hunter.
\newblock Matplotlib: A 2d graphics environment.
\newblock \emph{Computing in Science \& Engineering}, 9\penalty0 (3):\penalty0
  90--95, 2007.
\newblock \doi{10.1109/MCSE.2007.55}.

\bibitem[Jain et~al.(2021)Jain, Tancik, and Abbeel]{jain2021putting}
A.~Jain, M.~Tancik, and P.~Abbeel.
\newblock Putting nerf on a diet: Semantically consistent few-shot view
  synthesis.
\newblock In \emph{Proceedings of the IEEE/CVF International Conference on
  Computer Vision}, pages 5885--5894, 2021.

\bibitem[Jun and Nichol(2023)]{jun2023shap}
H.~Jun and A.~Nichol.
\newblock Shap-e: Generating conditional 3d implicit functions.
\newblock \emph{arXiv preprint arXiv:2305.02463}, 2023.

\bibitem[Kaplan et~al.(2020)Kaplan, McCandlish, Henighan, Brown, Chess, Child,
  Gray, Radford, Wu, and Amodei]{kaplan2020scaling}
J.~Kaplan, S.~McCandlish, T.~Henighan, T.~B. Brown, B.~Chess, R.~Child,
  S.~Gray, A.~Radford, J.~Wu, and D.~Amodei.
\newblock Scaling laws for neural language models.
\newblock \emph{arXiv preprint arXiv:2001.08361}, 2020.

\bibitem[Kato et~al.(2018)Kato, Ushiku, and Harada]{kato2018neural}
H.~Kato, Y.~Ushiku, and T.~Harada.
\newblock Neural 3d mesh renderer.
\newblock In \emph{Proceedings of the IEEE conference on computer vision and
  pattern recognition}, pages 3907--3916, 2018.

\bibitem[Kirillov et~al.(2023)Kirillov, Mintun, Ravi, Mao, Rolland, Gustafson,
  Xiao, Whitehead, Berg, Lo, et~al.]{kirillov2023segment}
A.~Kirillov, E.~Mintun, N.~Ravi, H.~Mao, C.~Rolland, L.~Gustafson, T.~Xiao,
  S.~Whitehead, A.~C. Berg, W.-Y. Lo, et~al.
\newblock Segment anything.
\newblock \emph{arXiv preprint arXiv:2304.02643}, 2023.

\bibitem[Lim et~al.(2013)Lim, Pirsiavash, and Torralba]{lim2013parsing}
J.~J. Lim, H.~Pirsiavash, and A.~Torralba.
\newblock Parsing ikea objects: Fine pose estimation.
\newblock In \emph{Proceedings of the IEEE international conference on computer
  vision}, pages 2992--2999, 2013.

\bibitem[Lin et~al.(2023)Lin, Gao, Tang, Takikawa, Zeng, Huang, Kreis, Fidler,
  Liu, and Lin]{lin2023magic3d}
C.-H. Lin, J.~Gao, L.~Tang, T.~Takikawa, X.~Zeng, X.~Huang, K.~Kreis,
  S.~Fidler, M.-Y. Liu, and T.-Y. Lin.
\newblock Magic3d: High-resolution text-to-3d content creation.
\newblock In \emph{Proceedings of the IEEE/CVF Conference on Computer Vision
  and Pattern Recognition}, pages 300--309, 2023.

\bibitem[Liu and Vondrick(2023)]{liu2023humans}
R.~Liu and C.~Vondrick.
\newblock Humans as light bulbs: 3d human reconstruction from thermal
  reflection.
\newblock In \emph{Proceedings of the IEEE/CVF Conference on Computer Vision
  and Pattern Recognition}, pages 12531--12542, 2023.

\bibitem[Liu et~al.(2022)Liu, Menon, Mao, Park, Stent, and
  Vondrick]{liu2022shadows}
R.~Liu, S.~Menon, C.~Mao, D.~Park, S.~Stent, and C.~Vondrick.
\newblock Shadows shed light on 3d objects.
\newblock \emph{arXiv preprint arXiv:2206.08990}, 2022.

\bibitem[Liu et~al.(2023)Liu, Wu, Hoorick, Tokmakov, Zakharov, and
  Vondrick]{liu2023zero1to3}
R.~Liu, R.~Wu, B.~V. Hoorick, P.~Tokmakov, S.~Zakharov, and C.~Vondrick.
\newblock Zero-1-to-3: Zero-shot one image to 3d object, 2023.

\bibitem[Lu et~al.(2022)Lu, Clark, Zellers, Mottaghi, and
  Kembhavi]{Lu2022UnifiedIOAU}
J.~Lu, C.~Clark, R.~Zellers, R.~Mottaghi, and A.~Kembhavi.
\newblock Unified-io: A unified model for vision, language, and multi-modal
  tasks.
\newblock \emph{ArXiv}, abs/2206.08916, 2022.

\bibitem[Mescheder et~al.(2019)Mescheder, Oechsle, Niemeyer, Nowozin, and
  Geiger]{mescheder2019occupancy}
L.~Mescheder, M.~Oechsle, M.~Niemeyer, S.~Nowozin, and A.~Geiger.
\newblock Occupancy networks: Learning 3d reconstruction in function space.
\newblock In \emph{Proceedings of the IEEE/CVF conference on computer vision
  and pattern recognition}, pages 4460--4470, 2019.

\bibitem[Mildenhall et~al.(2020)Mildenhall, Srinivasan, Tancik, Barron,
  Ramamoorthi, and Ng]{mildenhall2020nerf}
B.~Mildenhall, P.~P. Srinivasan, M.~Tancik, J.~T. Barron, R.~Ramamoorthi, and
  R.~Ng.
\newblock Nerf: Representing scenes as neural radiance fields for view
  synthesis.
\newblock In \emph{ECCV}, 2020.

\bibitem[Miller(1995)]{miller1995wordnet}
G.~A. Miller.
\newblock Wordnet: a lexical database for english.
\newblock \emph{Communications of the ACM}, 38\penalty0 (11):\penalty0 39--41,
  1995.

\bibitem[{Morrison} et~al.(2020){Morrison}, {Corke}, and
  {Leitner}]{morrison2020egad}
D.~{Morrison}, P.~{Corke}, and J.~{Leitner}.
\newblock Egad! an evolved grasping analysis dataset for diversity and
  reproducibility in robotic manipulation.
\newblock \emph{IEEE Robotics and Automation Letters}, 5\penalty0 (3):\penalty0
  4368--4375, 2020.

\bibitem[Nichol et~al.(2022)Nichol, Jun, Dhariwal, Mishkin, and
  Chen]{nichol2022point}
A.~Nichol, H.~Jun, P.~Dhariwal, P.~Mishkin, and M.~Chen.
\newblock Point-e: A system for generating 3d point clouds from complex
  prompts.
\newblock \emph{arXiv preprint arXiv:2212.08751}, 2022.

\bibitem[OpenAI(2023)]{openai2023gpt}
OpenAI.
\newblock Gpt-4 technical report.
\newblock \emph{arXiv}, 2023.

\bibitem[Ouyang et~al.(2022)Ouyang, Wu, Jiang, Almeida, Wainwright, Mishkin,
  Zhang, Agarwal, Slama, Ray, et~al.]{ouyang2022training}
L.~Ouyang, J.~Wu, X.~Jiang, D.~Almeida, C.~Wainwright, P.~Mishkin, C.~Zhang,
  S.~Agarwal, K.~Slama, A.~Ray, et~al.
\newblock Training language models to follow instructions with human feedback.
\newblock \emph{Advances in Neural Information Processing Systems},
  35:\penalty0 27730--27744, 2022.

\bibitem[pandas~development team(2020)]{reback2020pandas}
T.~pandas~development team.
\newblock pandas-dev/pandas: Pandas, Feb. 2020.
\newblock URL \url{https://doi.org/10.5281/zenodo.3509134}.

\bibitem[Park et~al.(2018)Park, Rematas, Farhadi, and
  Seitz]{park2018photoshape}
K.~Park, K.~Rematas, A.~Farhadi, and S.~M. Seitz.
\newblock Photoshape: Photorealistic materials for large-scale shape
  collections.
\newblock \emph{arXiv preprint arXiv:1809.09761}, 2018.

\bibitem[Paszke et~al.(2019)Paszke, Gross, Massa, Lerer, Bradbury, Chanan,
  Killeen, Lin, Gimelshein, Antiga, et~al.]{paszke2019pytorch}
A.~Paszke, S.~Gross, F.~Massa, A.~Lerer, J.~Bradbury, G.~Chanan, T.~Killeen,
  Z.~Lin, N.~Gimelshein, L.~Antiga, et~al.
\newblock Pytorch: An imperative style, high-performance deep learning library.
\newblock \emph{Advances in neural information processing systems}, 32, 2019.

\bibitem[Poole et~al.(2022)Poole, Jain, Barron, and
  Mildenhall]{poole2022dreamfusion}
B.~Poole, A.~Jain, J.~T. Barron, and B.~Mildenhall.
\newblock Dreamfusion: Text-to-3d using 2d diffusion.
\newblock \emph{arXiv preprint arXiv:2209.14988}, 2022.

\bibitem[Radford et~al.(2019)Radford, Wu, Child, Luan, Amodei, Sutskever,
  et~al.]{radford2019language}
A.~Radford, J.~Wu, R.~Child, D.~Luan, D.~Amodei, I.~Sutskever, et~al.
\newblock Language models are unsupervised multitask learners.
\newblock \emph{OpenAI blog}, 1\penalty0 (8):\penalty0 9, 2019.

\bibitem[Radford et~al.(2021)Radford, Kim, Hallacy, Ramesh, Goh, Agarwal,
  Sastry, Askell, Mishkin, Clark, et~al.]{radford2021learning}
A.~Radford, J.~W. Kim, C.~Hallacy, A.~Ramesh, G.~Goh, S.~Agarwal, G.~Sastry,
  A.~Askell, P.~Mishkin, J.~Clark, et~al.
\newblock Learning transferable visual models from natural language
  supervision.
\newblock In \emph{International conference on machine learning}, pages
  8748--8763. PMLR, 2021.

\bibitem[Ramesh et~al.(2022)Ramesh, Dhariwal, Nichol, Chu, and
  Chen]{ramesh2022hierarchical}
A.~Ramesh, P.~Dhariwal, A.~Nichol, C.~Chu, and M.~Chen.
\newblock Hierarchical text-conditional image generation with clip latents.
\newblock \emph{arXiv preprint arXiv:2204.06125}, 2022.

\bibitem[Ravi et~al.(2020)Ravi, Reizenstein, Novotny, Gordon, Lo, Johnson, and
  Gkioxari]{ravi2020accelerating}
N.~Ravi, J.~Reizenstein, D.~Novotny, T.~Gordon, W.-Y. Lo, J.~Johnson, and
  G.~Gkioxari.
\newblock Accelerating 3d deep learning with pytorch3d.
\newblock \emph{arXiv preprint arXiv:2007.08501}, 2020.

\bibitem[Ren et~al.(2015)Ren, He, Girshick, and Sun]{ren2015faster}
S.~Ren, K.~He, R.~Girshick, and J.~Sun.
\newblock Faster r-cnn: Towards real-time object detection with region proposal
  networks.
\newblock \emph{Advances in neural information processing systems}, 28, 2015.

\bibitem[Rombach et~al.(2022)Rombach, Blattmann, Lorenz, Esser, and
  Ommer]{rombach2022high}
R.~Rombach, A.~Blattmann, D.~Lorenz, P.~Esser, and B.~Ommer.
\newblock High-resolution image synthesis with latent diffusion models.
\newblock In \emph{Proceedings of the IEEE/CVF Conference on Computer Vision
  and Pattern Recognition}, pages 10684--10695, 2022.

\bibitem[Schuhmann et~al.(2022)Schuhmann, Beaumont, Vencu, Gordon, Wightman,
  Cherti, Coombes, Katta, Mullis, Wortsman, et~al.]{schuhmann2022laion}
C.~Schuhmann, R.~Beaumont, R.~Vencu, C.~Gordon, R.~Wightman, M.~Cherti,
  T.~Coombes, A.~Katta, C.~Mullis, M.~Wortsman, et~al.
\newblock Laion-5b: An open large-scale dataset for training next generation
  image-text models.
\newblock \emph{arXiv preprint arXiv:2210.08402}, 2022.

\bibitem[Tang(2022)]{stable-dreamfusion}
J.~Tang.
\newblock Stable-dreamfusion: Text-to-3d with stable-diffusion, 2022.
\newblock https://github.com/ashawkey/stable-dreamfusion.

\bibitem[Touvron et~al.(2023)Touvron, Lavril, Izacard, Martinet, Lachaux,
  Lacroix, Rozi{\`e}re, Goyal, Hambro, Azhar, et~al.]{touvron2023llama}
H.~Touvron, T.~Lavril, G.~Izacard, X.~Martinet, M.-A. Lachaux, T.~Lacroix,
  B.~Rozi{\`e}re, N.~Goyal, E.~Hambro, F.~Azhar, et~al.
\newblock Llama: Open and efficient foundation language models.
\newblock \emph{arXiv preprint arXiv:2302.13971}, 2023.

\bibitem[Wang et~al.(2023)Wang, Du, Li, Yeh, and Shakhnarovich]{wang2023score}
H.~Wang, X.~Du, J.~Li, R.~A. Yeh, and G.~Shakhnarovich.
\newblock Score jacobian chaining: Lifting pretrained 2d diffusion models for
  3d generation.
\newblock In \emph{Proceedings of the IEEE/CVF Conference on Computer Vision
  and Pattern Recognition}, pages 12619--12629, 2023.

\bibitem[Wang et~al.(2018)Wang, Zhang, Li, Fu, Liu, and
  Jiang]{wang2018pixel2mesh}
N.~Wang, Y.~Zhang, Z.~Li, Y.~Fu, W.~Liu, and Y.-G. Jiang.
\newblock Pixel2mesh: Generating 3d mesh models from single rgb images.
\newblock In \emph{Proceedings of the European conference on computer vision
  (ECCV)}, pages 52--67, 2018.

\bibitem[Wang et~al.(2021)Wang, Wang, Genova, Srinivasan, Zhou, Barron,
  Martin-Brualla, Snavely, and Funkhouser]{wang2021ibrnet}
Q.~Wang, Z.~Wang, K.~Genova, P.~P. Srinivasan, H.~Zhou, J.~T. Barron,
  R.~Martin-Brualla, N.~Snavely, and T.~Funkhouser.
\newblock Ibrnet: Learning multi-view image-based rendering.
\newblock In \emph{Proceedings of the IEEE/CVF Conference on Computer Vision
  and Pattern Recognition}, pages 4690--4699, 2021.

\bibitem[Waskom(2021)]{Waskom2021}
M.~L. Waskom.
\newblock seaborn: statistical data visualization.
\newblock \emph{Journal of Open Source Software}, 6\penalty0 (60):\penalty0
  3021, 2021.
\newblock \doi{10.21105/joss.03021}.
\newblock URL \url{https://doi.org/10.21105/joss.03021}.

\bibitem[Wu et~al.(2023{\natexlab{a}})Wu, Johnson, Malik, Feichtenhofer, and
  Gkioxari]{wu2023multiview}
C.-Y. Wu, J.~Johnson, J.~Malik, C.~Feichtenhofer, and G.~Gkioxari.
\newblock Multiview compressive coding for 3d reconstruction.
\newblock \emph{arXiv preprint arXiv:2301.08247}, 2023{\natexlab{a}}.

\bibitem[Wu et~al.(2023{\natexlab{b}})Wu, Zhang, Fu, Wang, Ren, Pan, Wu, Yang,
  Wang, Qian, Lin, and Liu]{wu2023omniobject3d}
T.~Wu, J.~Zhang, X.~Fu, Y.~Wang, J.~Ren, L.~Pan, W.~Wu, L.~Yang, J.~Wang,
  C.~Qian, D.~Lin, and Z.~Liu.
\newblock Omniobject3d: Large-vocabulary 3d object dataset for realistic
  perception, reconstruction and generation.
\newblock \emph{IEEE/CVF Conference on Computer Vision and Pattern Recognition
  (CVPR)}, 2023{\natexlab{b}}.

\bibitem[Yu et~al.(2021)Yu, Ye, Tancik, and Kanazawa]{yu2021pixelnerf}
A.~Yu, V.~Ye, M.~Tancik, and A.~Kanazawa.
\newblock pixelnerf: Neural radiance fields from one or few images.
\newblock In \emph{Proceedings of the IEEE/CVF Conference on Computer Vision
  and Pattern Recognition}, pages 4578--4587, 2021.

\bibitem[Zhang et~al.(2018)Zhang, Isola, Efros, Shechtman, and
  Wang]{zhang2018unreasonable}
R.~Zhang, P.~Isola, A.~A. Efros, E.~Shechtman, and O.~Wang.
\newblock The unreasonable effectiveness of deep features as a perceptual
  metric.
\newblock In \emph{Proceedings of the IEEE conference on computer vision and
  pattern recognition}, pages 586--595, 2018.

\bibitem[Zhou et~al.(2023)Zhou, Liu, Xu, Iyer, Sun, Mao, Ma, Efrat, Yu, Yu,
  et~al.]{zhou2023lima}
C.~Zhou, P.~Liu, P.~Xu, S.~Iyer, J.~Sun, Y.~Mao, X.~Ma, A.~Efrat, P.~Yu, L.~Yu,
  et~al.
\newblock Lima: Less is more for alignment.
\newblock \emph{arXiv preprint arXiv:2305.11206}, 2023.

\bibitem[Zhou and Jacobson(2016)]{zhou2016thingi10k}
Q.~Zhou and A.~Jacobson.
\newblock Thingi10k: A dataset of 10,000 3d-printing models.
\newblock \emph{arXiv preprint arXiv:1605.04797}, 2016.

\end{thebibliography}

\appendix

\newpage

\section{Implementation Details}
\label{sec:appendixImpl}

\subsection{Zero123-XL}

A batch size of 2048 is used during training with a learning rate of 1e-4. Different from the original paper~\cite{liu2023zero1to3}, we performed a second-stage finetuning with a smaller learning rate of 5e-5 on a high-quality subset of Objaverse-XL selected with dataset metadata. The first stage was trained for 375K iterations and the second stage is trained for 65K iterations. For dataset scaling experiment whose results are shown in~\ref{fig:scale}, datasets with size below 800K are randomly sampled subsets from Objaverse 1.0. We keep the rest of the setting consistent with the original paper~\cite{liu2023zero1to3}. For calculating LPIPS metric in Figure~\ref{fig:scale}, we multiple the score by 10 for better visualization.

\section{Additional Zero123-XL Comparisons}

Figures~\ref{fig:aex1}-\ref{fig:aex12} show additional comparisons between Zero123-XL and Zero123. Overall, Zero123-XL shows better generalization than Zero123 by both better following the camera transformation and generating more plausible outputs.

\begin{figure}
    \centering
    \begin{adjustbox}{center}
        \includegraphics[width=0.97\textwidth]{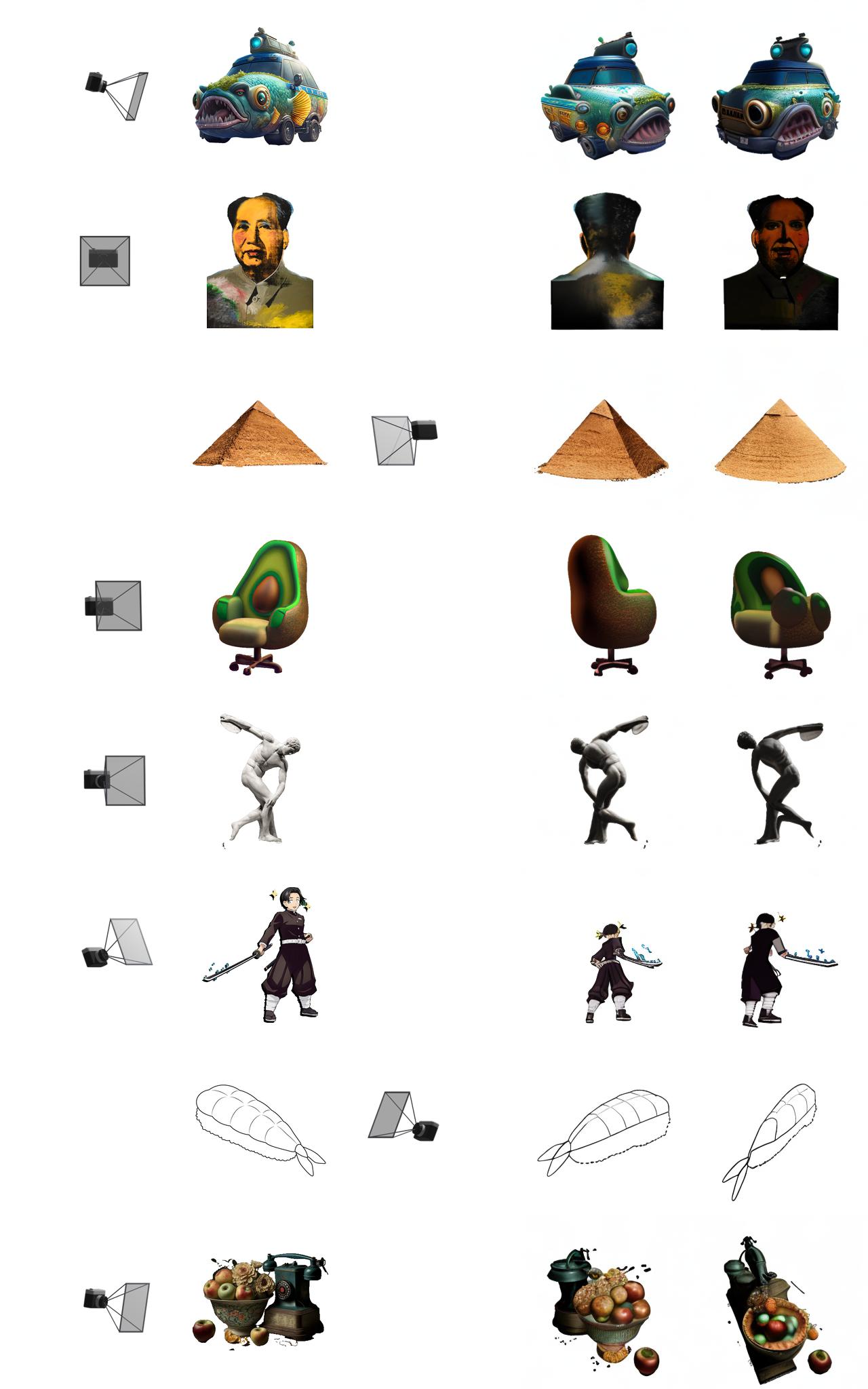}
    \end{adjustbox}
    \begin{adjustbox}{center}
        \includegraphics[width=0.97\textwidth]{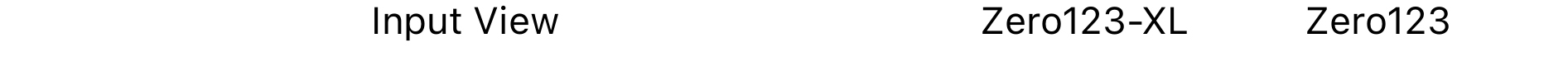}
    \end{adjustbox}
    \caption{Additional examples comparing the outputs of Zero123-XL and Zero123 under different camera transformations.}
    \label{fig:aex1}
\end{figure}

\begin{figure}
    \centering
    \begin{adjustbox}{center}
        \includegraphics[width=0.97\textwidth]{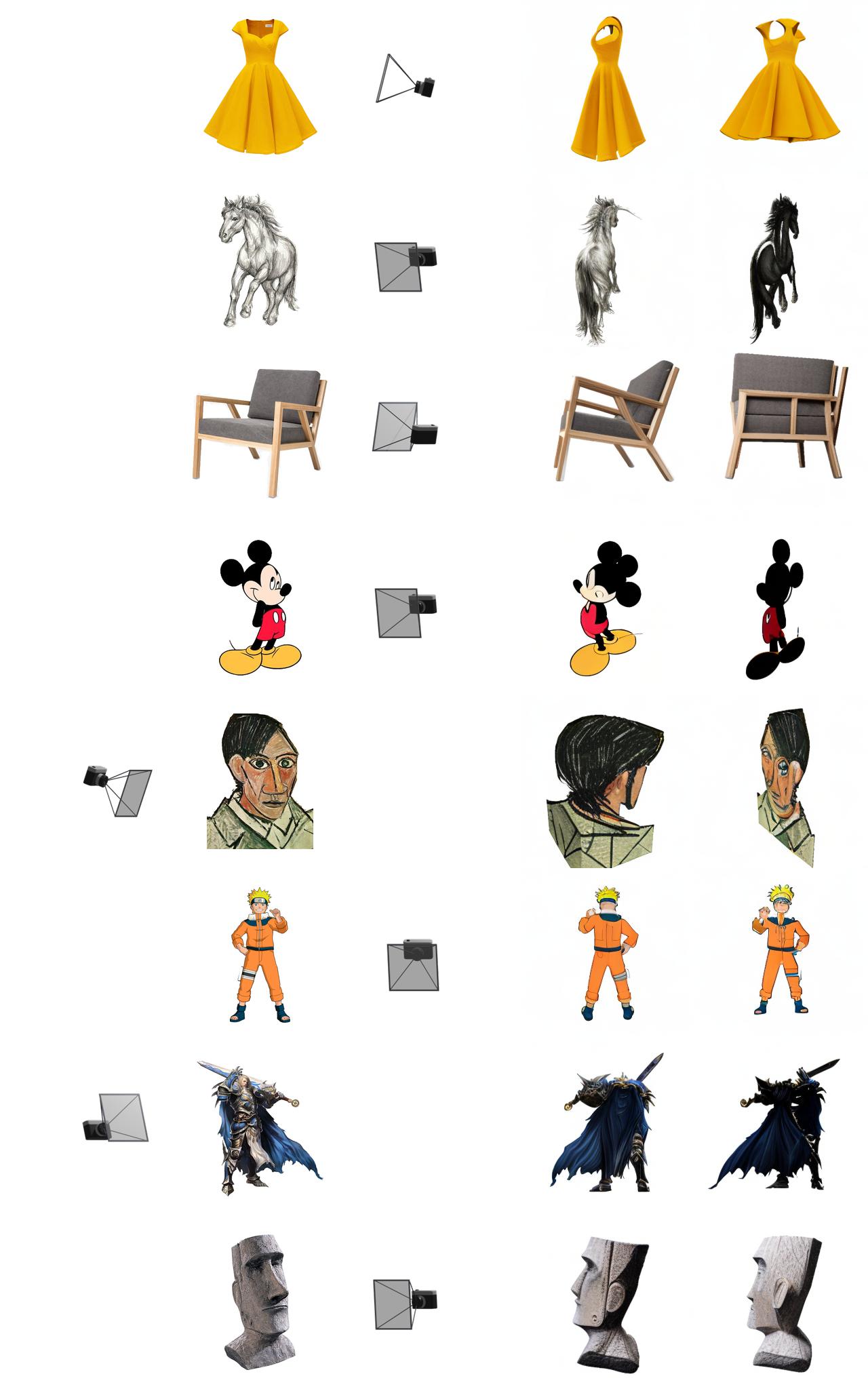}
    \end{adjustbox}
    \begin{adjustbox}{center}
        \includegraphics[width=0.97\textwidth]{images/camera-context.pdf}
    \end{adjustbox}
    \caption{Continuation of additional examples comparing Zero123-XL and Zero123.}
    \label{fig:aex2}
\end{figure}

\begin{figure}
    \centering
    \begin{adjustbox}{center}
        \includegraphics[width=0.97\textwidth]{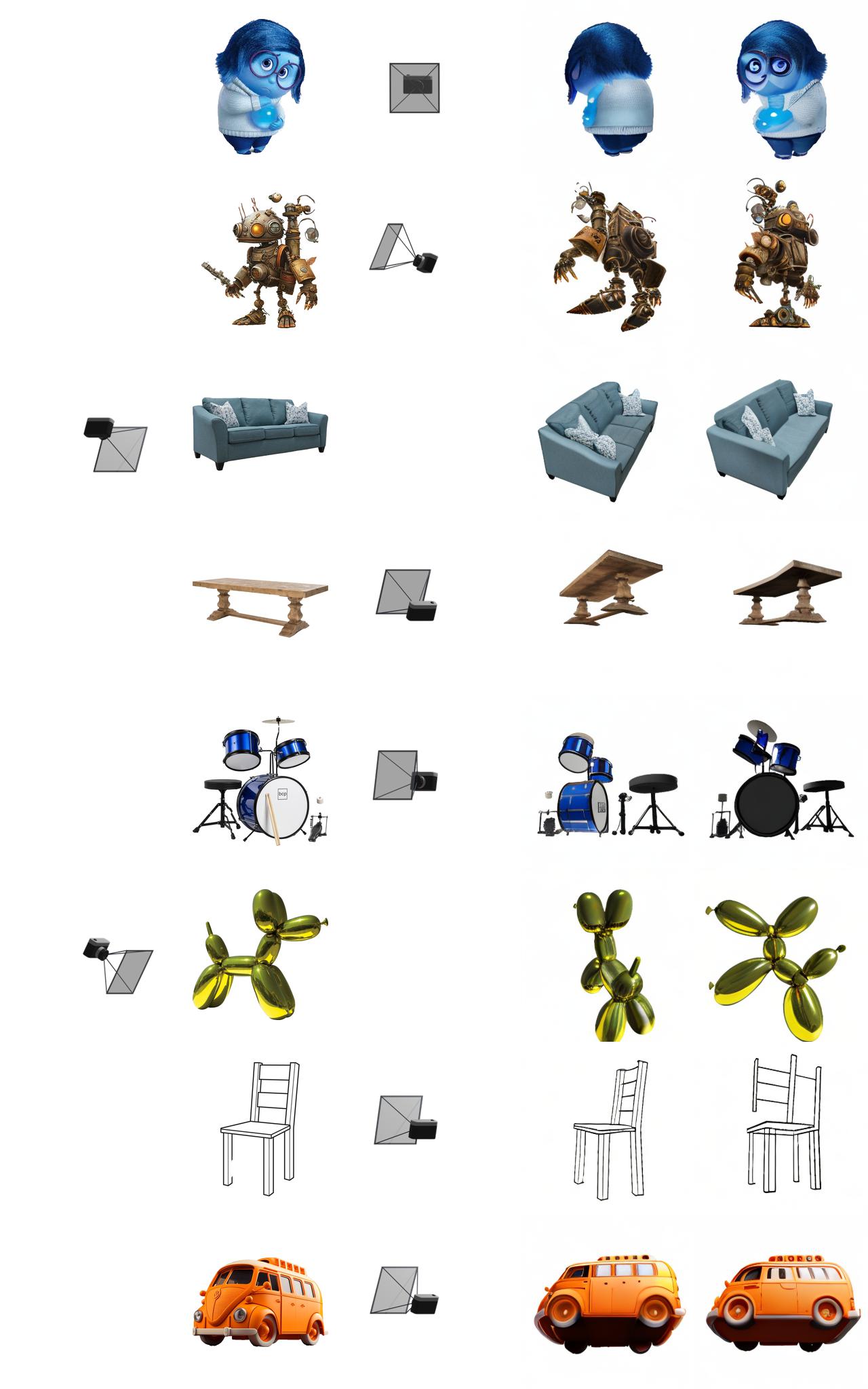}
    \end{adjustbox}
    \begin{adjustbox}{center}
        \includegraphics[width=0.97\textwidth]{images/camera-context.pdf}
    \end{adjustbox}
    \caption{Continuation of additional examples comparing Zero123-XL and Zero123.}
    \label{fig:aex3}
\end{figure}

\begin{figure}
    \centering
    \begin{adjustbox}{center}
        \includegraphics[width=0.97\textwidth]{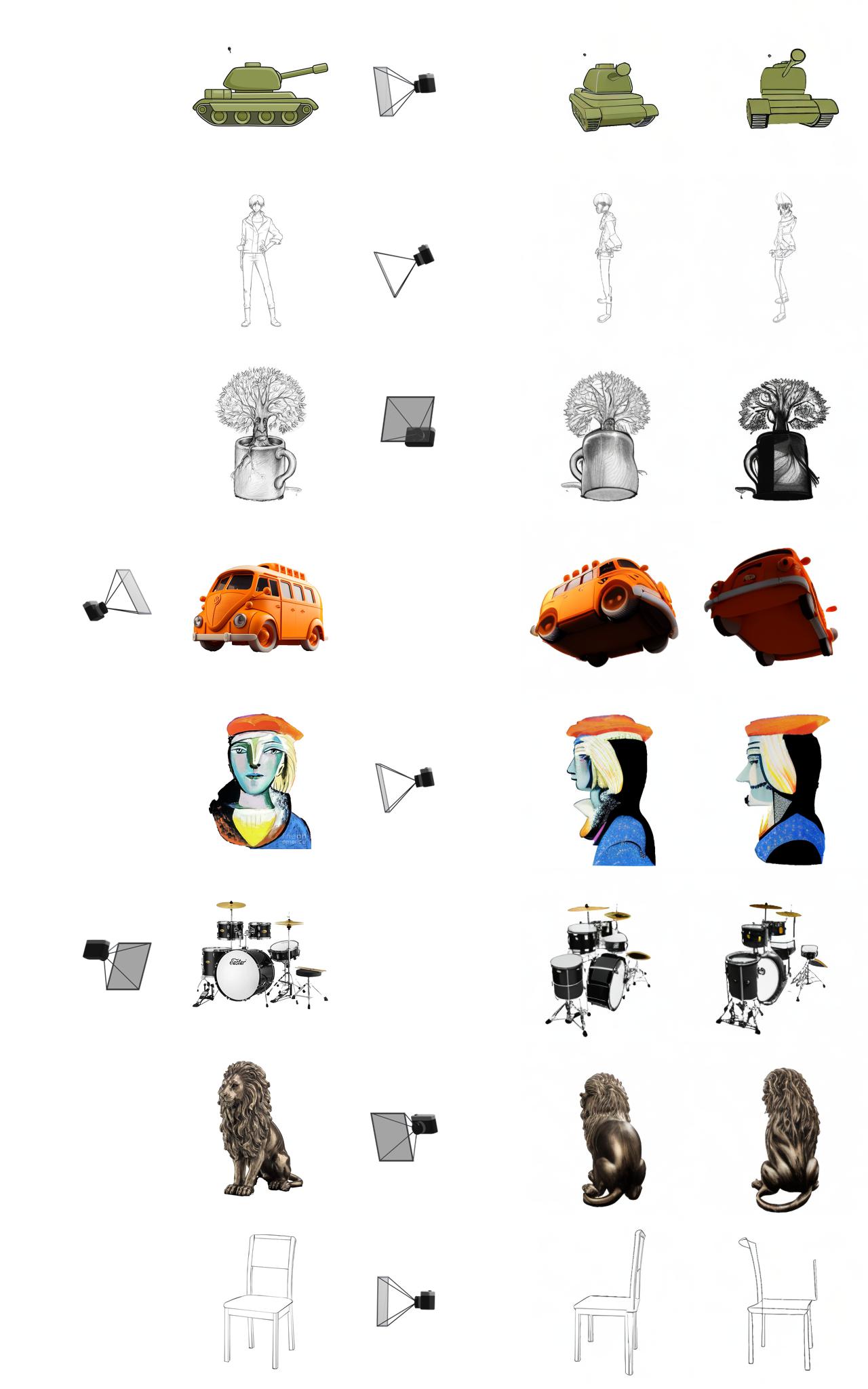}
    \end{adjustbox}
    \begin{adjustbox}{center}
        \includegraphics[width=0.97\textwidth]{images/camera-context.pdf}
    \end{adjustbox}
    \caption{Continuation of additional examples comparing Zero123-XL and Zero123.}
    \label{fig:aex4}
\end{figure}

\begin{figure}
    \centering
    \begin{adjustbox}{center}
        \includegraphics[width=0.97\textwidth]{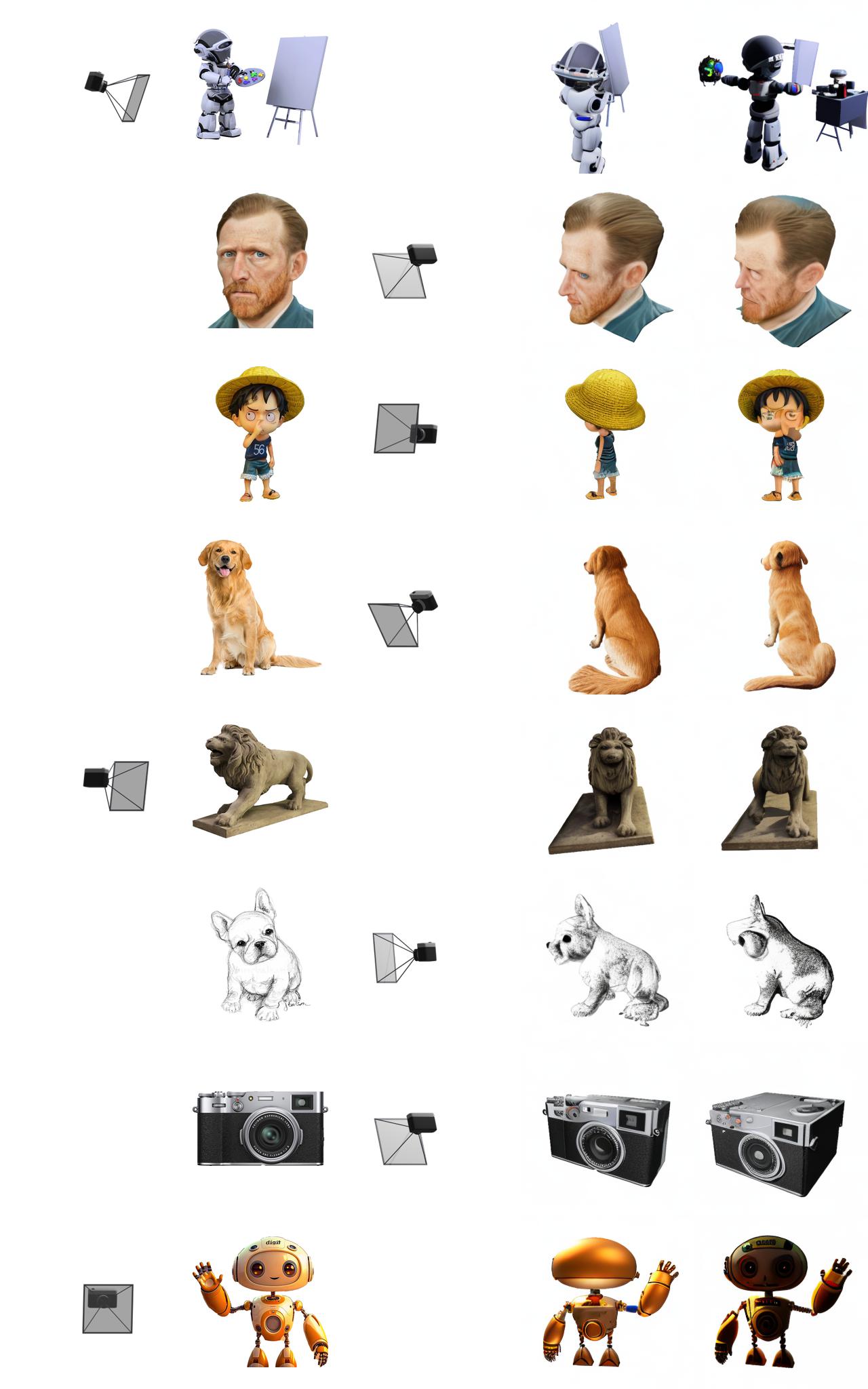}
    \end{adjustbox}
    \begin{adjustbox}{center}
        \includegraphics[width=0.97\textwidth]{images/camera-context.pdf}
    \end{adjustbox}
    \caption{Continuation of additional examples comparing Zero123-XL and Zero123.}
    \label{fig:aex5}
\end{figure}

\begin{figure}
    \centering
    \begin{adjustbox}{center}
        \includegraphics[width=0.97\textwidth]{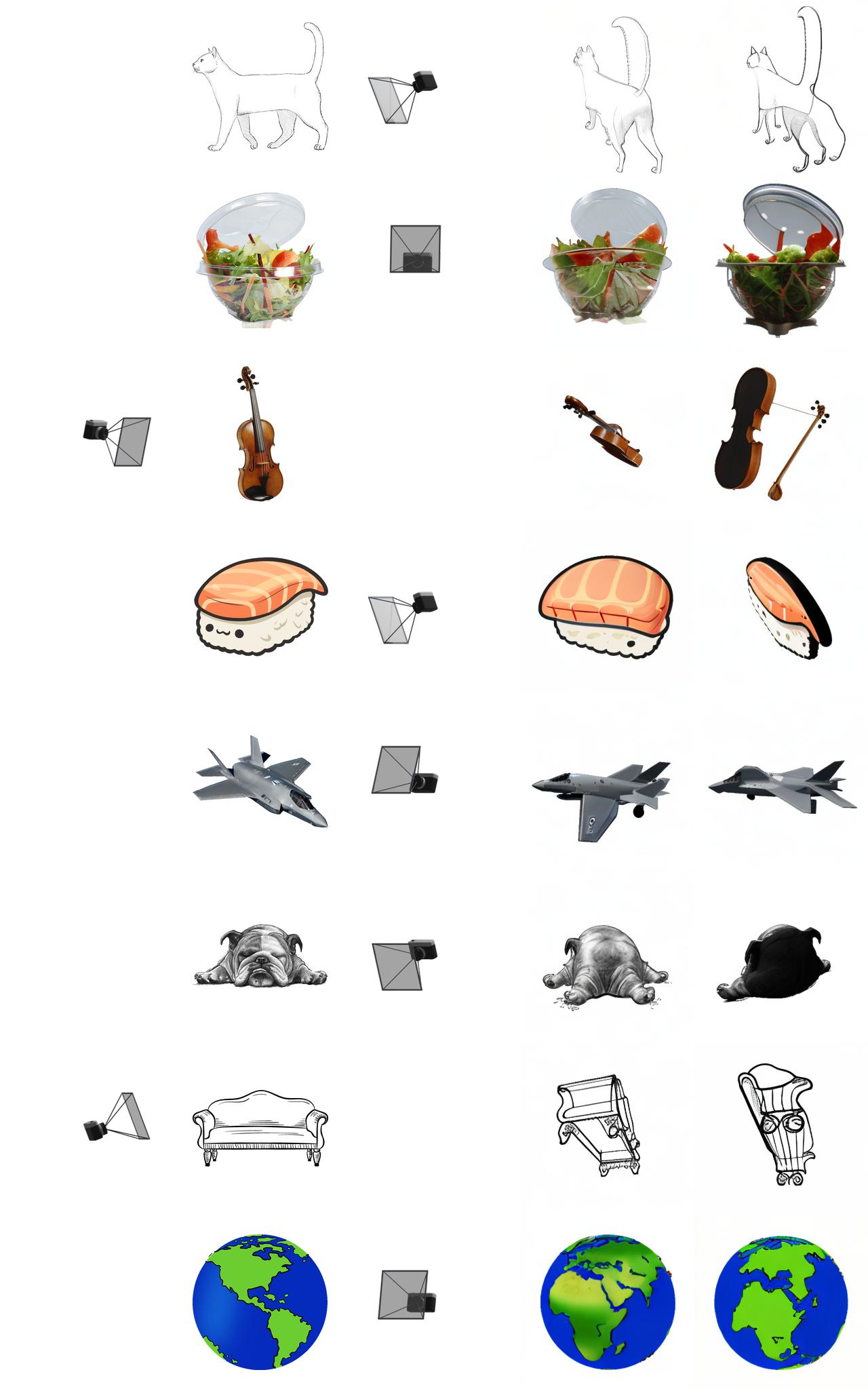}
    \end{adjustbox}
    \begin{adjustbox}{center}
        \includegraphics[width=0.97\textwidth]{images/camera-context.pdf}
    \end{adjustbox}
    \caption{Continuation of additional examples comparing Zero123-XL and Zero123.}
    \label{fig:aex6}
\end{figure}

\begin{figure}
    \centering
    \begin{adjustbox}{center}
        \includegraphics[width=0.97\textwidth]{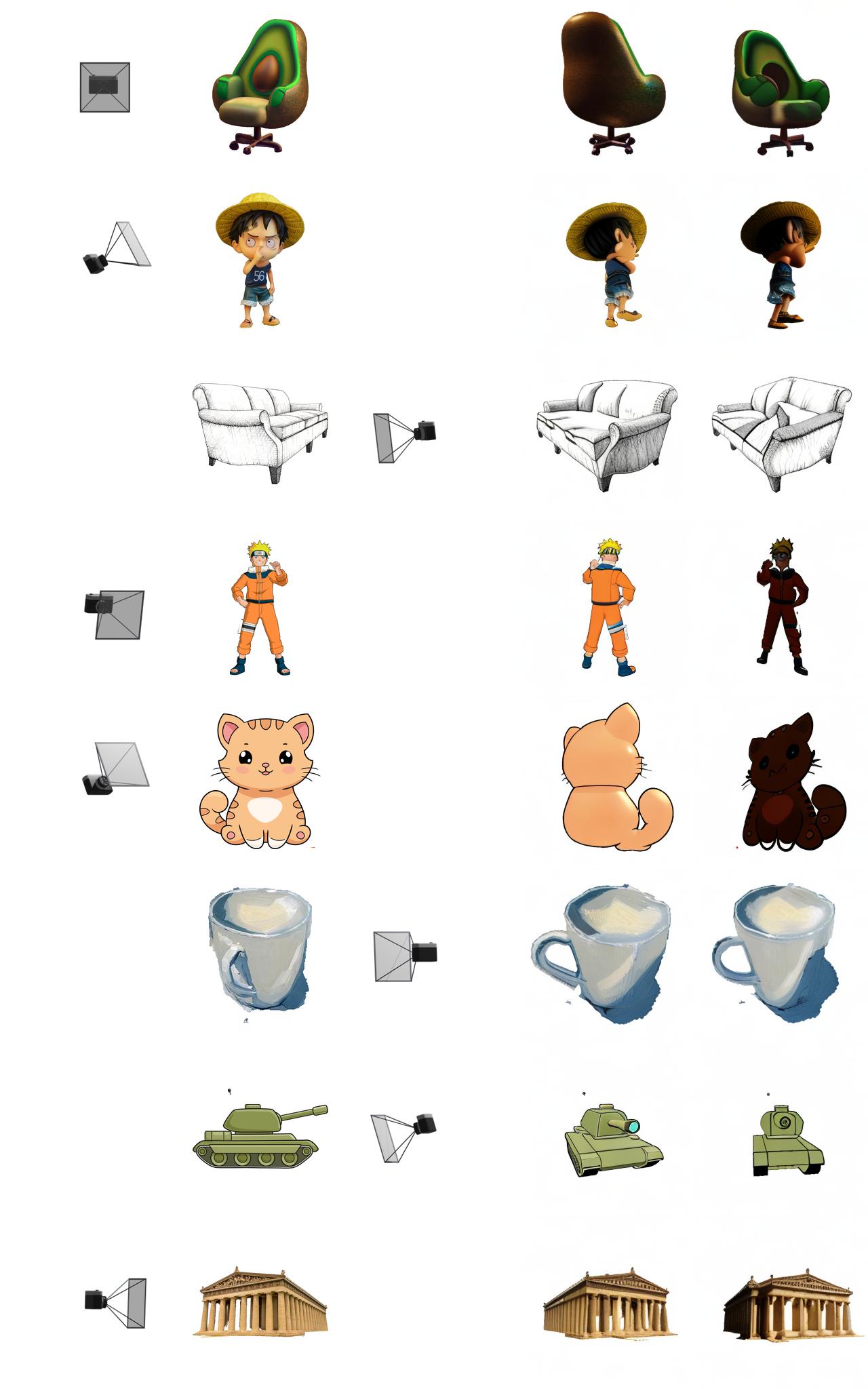}
    \end{adjustbox}
    \begin{adjustbox}{center}
        \includegraphics[width=0.97\textwidth]{images/camera-context.pdf}
    \end{adjustbox}
    \caption{Continuation of additional examples comparing Zero123-XL and Zero123.}
    \label{fig:aex7}
\end{figure}

\begin{figure}
    \centering
    \begin{adjustbox}{center}
        \includegraphics[width=0.97\textwidth]{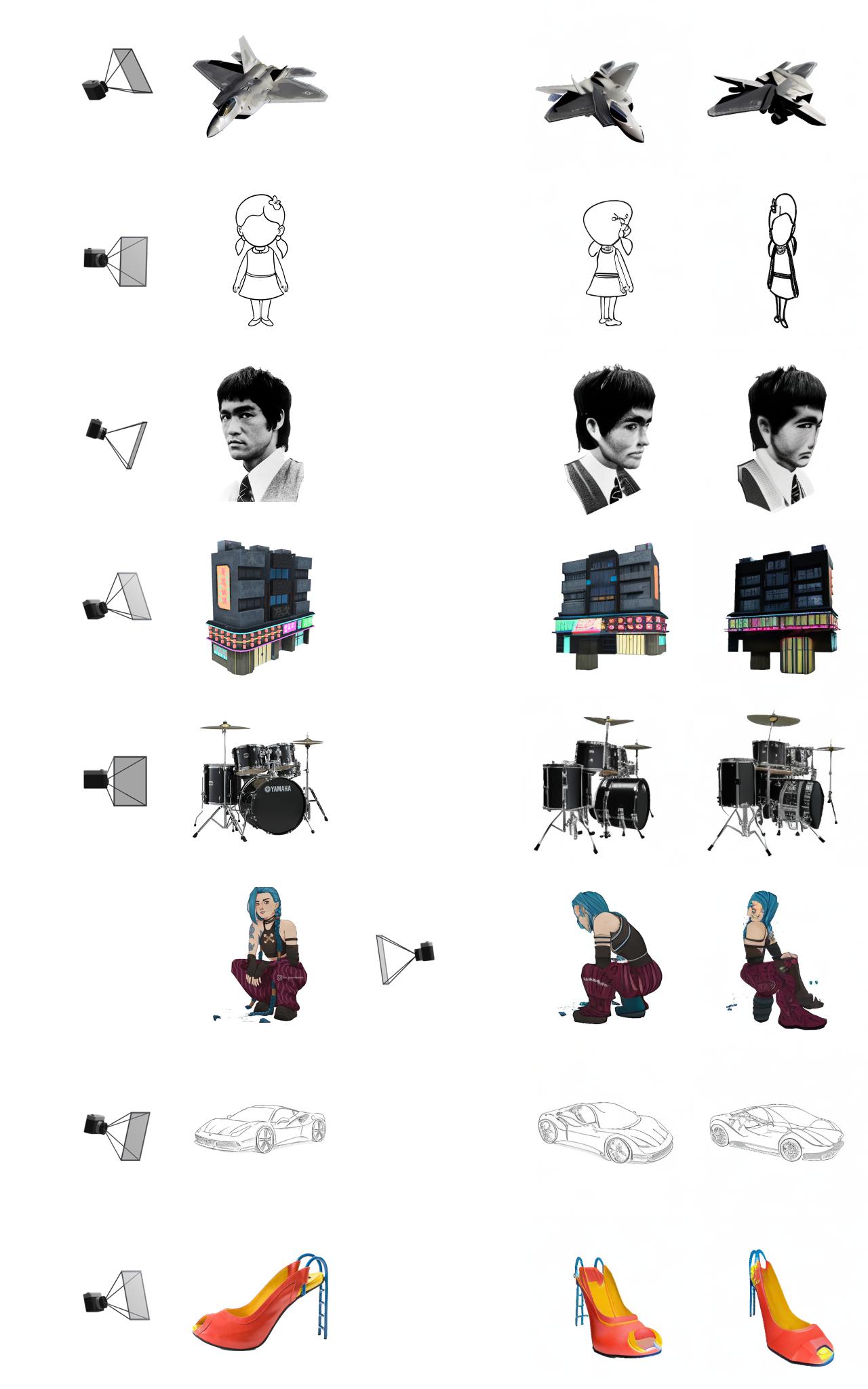}
    \end{adjustbox}
    \begin{adjustbox}{center}
        \includegraphics[width=0.97\textwidth]{images/camera-context.pdf}
    \end{adjustbox}
    \caption{Continuation of additional examples comparing Zero123-XL and Zero123.}
    \label{fig:aex8}
\end{figure}

\begin{figure}
    \centering
    \begin{adjustbox}{center}
        \includegraphics[width=0.97\textwidth]{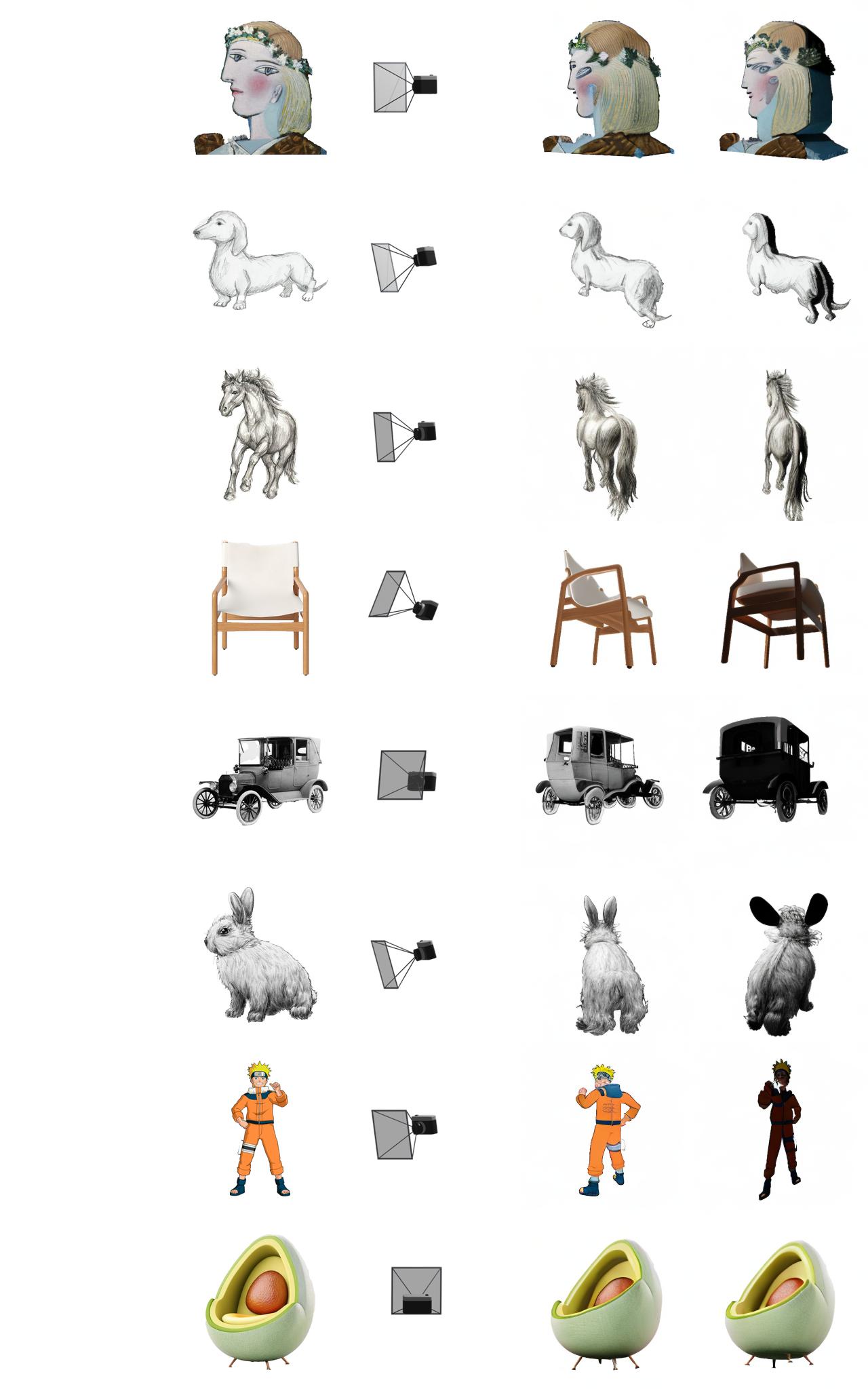}
    \end{adjustbox}
    \begin{adjustbox}{center}
        \includegraphics[width=0.97\textwidth]{images/camera-context.pdf}
    \end{adjustbox}
    \caption{Continuation of additional examples comparing Zero123-XL and Zero123.}
    \label{fig:aex9}
\end{figure}

\begin{figure}
    \centering
    \begin{adjustbox}{center}
        \includegraphics[width=0.97\textwidth]{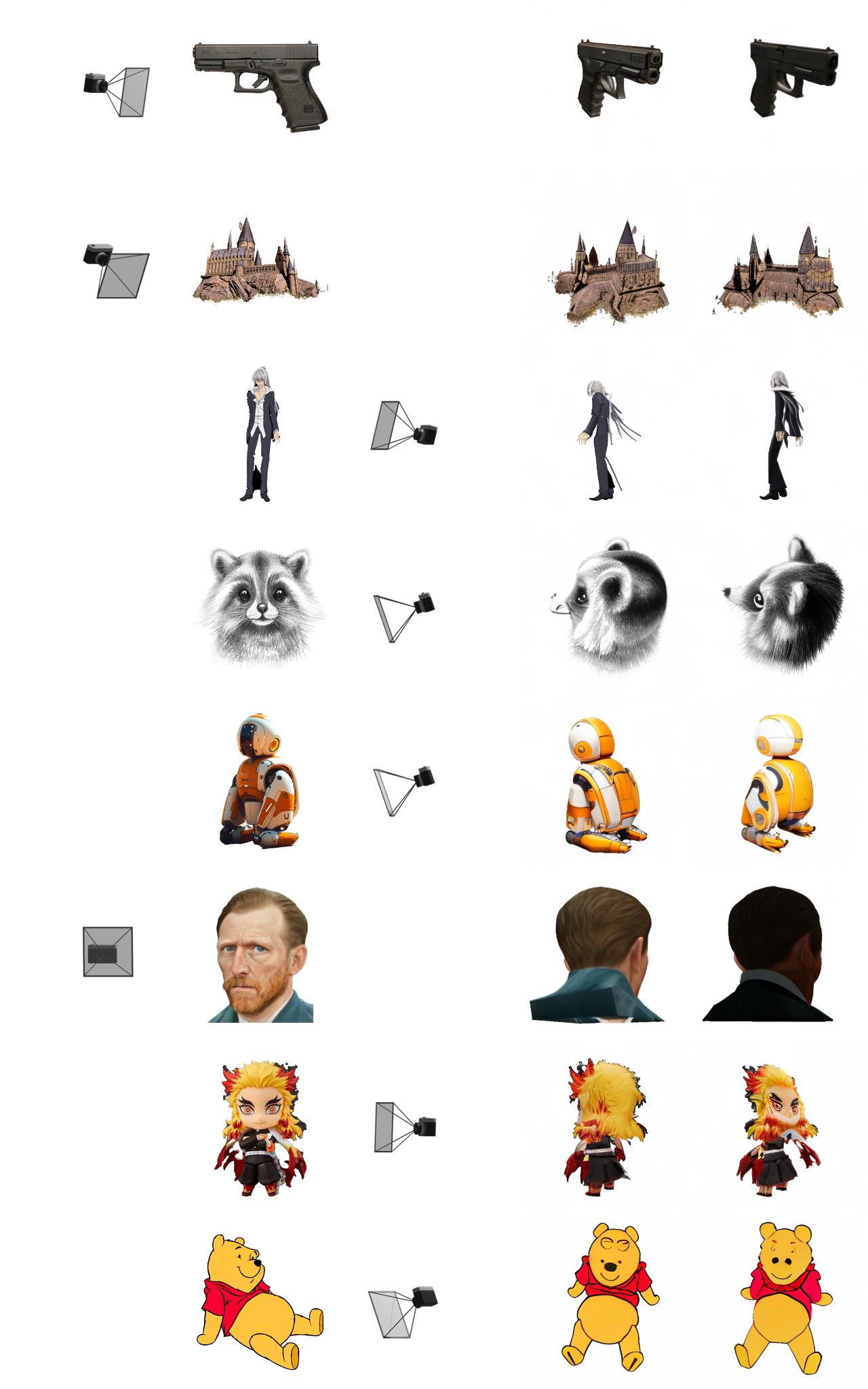}
    \end{adjustbox}
    \begin{adjustbox}{center}
        \includegraphics[width=0.97\textwidth]{images/camera-context.pdf}
    \end{adjustbox}
    \caption{Continuation of additional examples comparing Zero123-XL and Zero123.}
    \label{fig:aex10}
\end{figure}

\begin{figure}
    \centering
    \begin{adjustbox}{center}
        \includegraphics[width=0.97\textwidth]{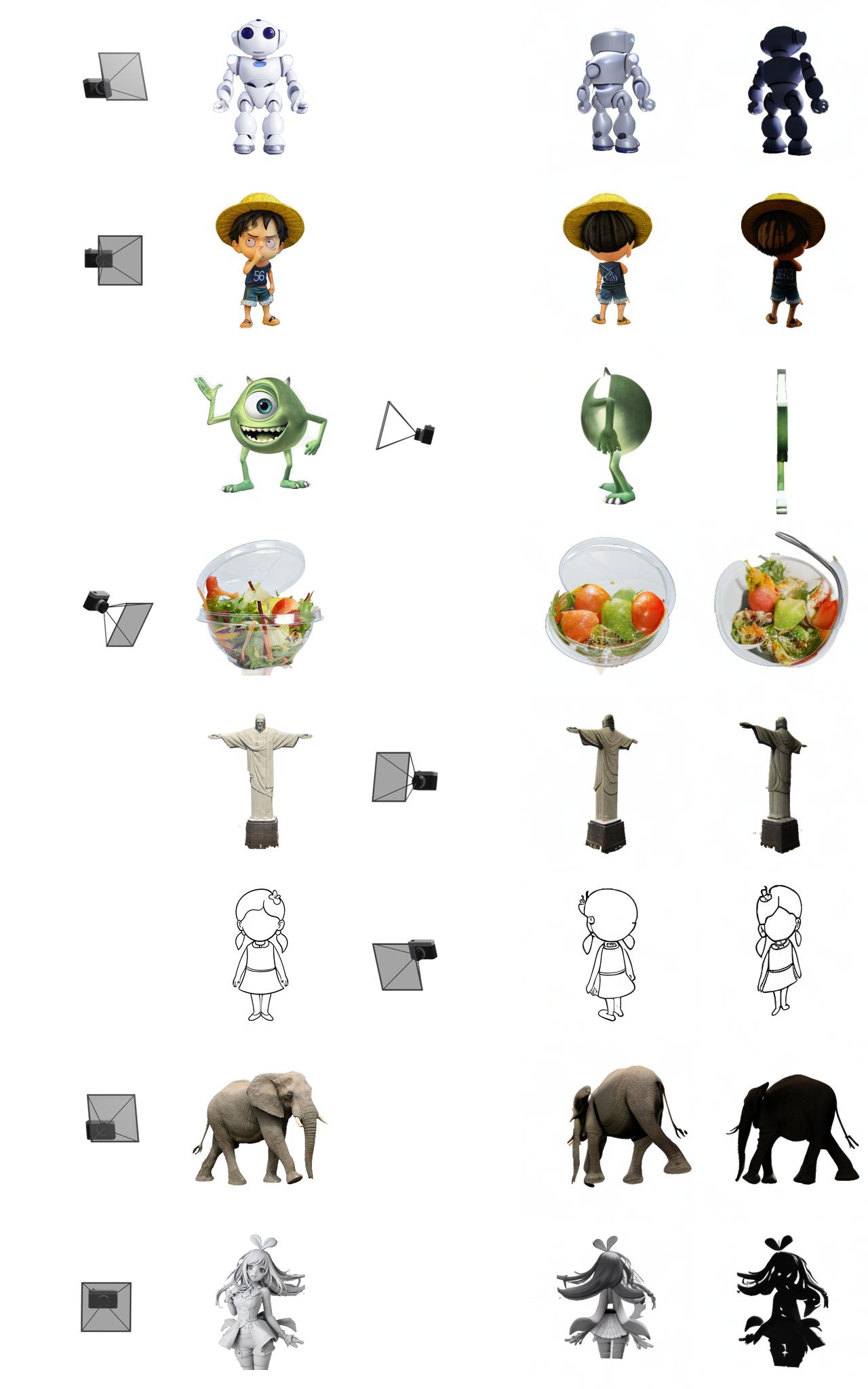}
    \end{adjustbox}
    \begin{adjustbox}{center}
        \includegraphics[width=0.97\textwidth]{images/camera-context.pdf}
    \end{adjustbox}
    \caption{Continuation of additional examples comparing Zero123-XL and Zero123.}
    \label{fig:aex11}
\end{figure}

\begin{figure}
    \centering
    \begin{adjustbox}{center}
        \includegraphics[width=0.97\textwidth]{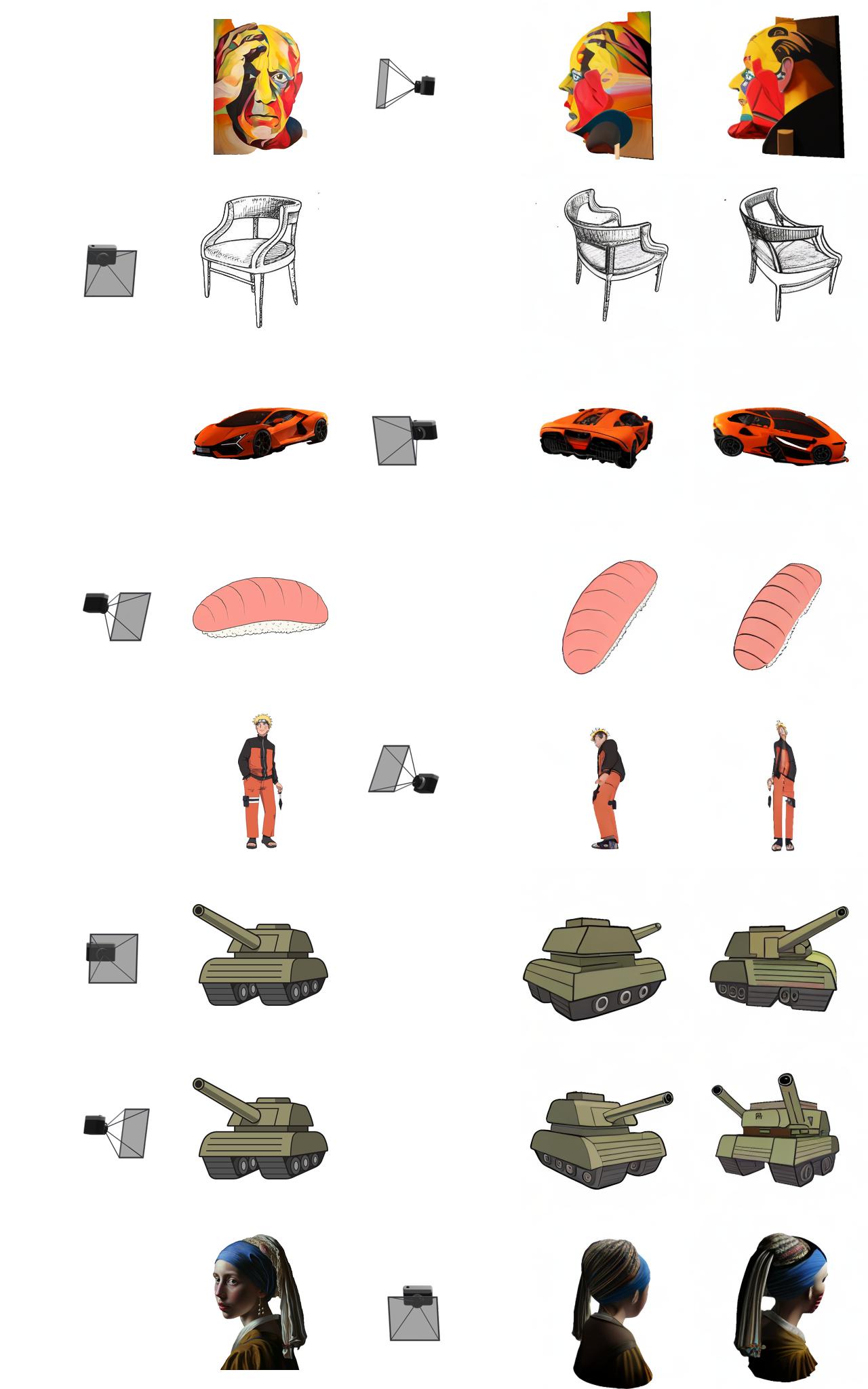}
    \end{adjustbox}
    \begin{adjustbox}{center}
        \includegraphics[width=0.97\textwidth]{images/camera-context.pdf}
    \end{adjustbox}
    \caption{Continuation of additional examples comparing Zero123-XL and Zero123.}
    \label{fig:aex12}
\end{figure}

\newpage

\section{Datasheet}
\label{sec:datasheet}

This section provides a datasheet~\cite{gebru2021datasheets} for \dataset.

\subsection{Motivation}

    \textcolor{\sectioncolor}{\textbf{
    For what purpose was the dataset created?
    }
    Was there a specific task in mind? Was there
    a specific gap that needed to be filled? Please provide a description.
    } \\
    The Objaverse-XL dataset was created to address the lack of high-quality, large-scale datasets for 3D vision tasks. This was due to challenges in acquiring such data and the associated complexities of 3D object generation and reconstruction, which were largely reliant on smaller, handcrafted datasets. The dataset was designed with the aim of advancing the field of 3D vision, allowing for the development and generalization improvement of models like Zero123 which work on tasks like novel view synthesis. The creation of Objaverse-XL essentially fills the gap in data availability for 3D vision tasks, particularly in light of increasing demand and interest in simulation, AR and VR technologies, and generative AI.
    \\
    
    \textcolor{\sectioncolor}{\textbf{
    Who created this dataset (e.g., which team, research group) and on behalf
    of which entity (e.g., company, institution, organization)?
    }
    } \\
    The dataset was created by researchers at the Allen Institute for AI and at the University of Washington, Seattle.
    \\
    
    \textcolor{\sectioncolor}{\textbf{
    What support was needed to make this dataset?
    }
    (e.g., who funded the creation of the dataset? If there is an associated grant, provide the name of the grantor and the grant name and number, or if it was supported by a company or government agency, give those details.)
    } \\
    Stability AI provided compute support and guidance for the main experiments in the paper. The Allen Institute for AI also provided compute support for collecting the dataset and performing rendering.
    \\
    
    \textcolor{\sectioncolor}{\textbf{
    Any other comments?
    }} \\
    No

\subsection{Composition}
    \textcolor{\sectioncolor}{\textbf{
    What do the instances that comprise the dataset represent (e.g., documents,
    photos, people, countries)?
    }
    Are there multiple types of instances (e.g., movies, users, and ratings;
    people and interactions between them; nodes and edges)? Please provide a
    description.
    } \\
    Instances of the dataset comprise of 3D objects and their associated metadata.
    \\
    
    \textcolor{\sectioncolor}{\textbf{
    How many instances are there in total (of each type, if appropriate)?
    }
    } \\
    There are approximately 10.2 million rendered 3D files. About 56\% come from GitHub, 35\% come from Thingiverse, 8\% come from Sketchfab, and less than 1\% come from Polycam and the Smithsonian Institute. We also release additional links to indexed GitHub files that are not included in the count due to being removed by deduplication or not being easily importable into Blender.\\
    
    \textcolor{\sectioncolor}{\textbf{
    Does the dataset contain all possible instances or is it a sample (not
    necessarily random) of instances from a larger set?
    }
    If the dataset is a sample, then what is the larger set? Is the sample
    representative of the larger set (e.g., geographic coverage)? If so, please
    describe how this representativeness was validated/verified. If it is not
    representative of the larger set, please describe why not (e.g., to cover a
    more diverse range of instances, because instances were withheld or
    unavailable).
    } \\
    The dataset contains a sample of objects on GitHub, Sketchfab, Thingiverse, and Polycam, along with all the objects from the Smithsonian Institute.
    \\
    
    \textcolor{\sectioncolor}{\textbf{
    What data does each instance consist of?
    }
    “Raw” data (e.g., unprocessed text or images) or features? In either case,
    please provide a description.
    } \\
    Instances of the dataset vary based on source. For Polycam and Sketchfab objects, we release the full 3D objects along with associated metadata. For GitHub, Thingiverse, and Smithsonian objects, we release links to each of the files that can then be downloaded from the source, along with metadata such as license, poly count, vertex count, and other attributes discussed in Section~\ref{sec:metadata}.
    \\
    
    \textcolor{\sectioncolor}{\textbf{
    Is there a label or target associated with each instance?
    }
    If so, please provide a description.
    } \\
    No, just the 3D object and associated metadata. Labels and targets may be derived from the metadata, but vary based on the task.\\
    
    \textcolor{\sectioncolor}{\textbf{
    Is any information missing from individual instances?
    }
    If so, please provide a description, explaining why this information is
    missing (e.g., because it was unavailable). This does not include
    intentionally removed information, but might include, e.g., redacted text.
    } \\
    No, information is not missing.\\
    
    \textcolor{\sectioncolor}{\textbf{
    Are relationships between individual instances made explicit (e.g., users’
    movie ratings, social network links)?
    }
    If so, please describe how these relationships are made explicit.
    } \\
    Individual instances are treated as independent. \\
    
    \textcolor{\sectioncolor}{\textbf{
    Are there recommended data splits (e.g., training, development/validation,
    testing)?
    }
    If so, please provide a description of these splits, explaining the
    rationale behind them.
    } \\
    There are no recommended data splits across the entire dataset as splits vary based on task.\\
    
    \textcolor{\sectioncolor}{\textbf{
    Are there any errors, sources of noise, or redundancies in the dataset?
    }
    If so, please provide a description.
    } \\
    We deduplicated the objects by taking a sha256 of the file contents. There may still be near duplicates that exist, if the objects are slightly modified, which could potentially be filtered out, if desirable, using CLIP embeddings of the renders.\\
    
    \textcolor{\sectioncolor}{\textbf{
    Is the dataset self-contained, or does it link to or otherwise rely on
    external resources (e.g., websites, tweets, other datasets)?
    }
    If it links to or relies on external resources, a) are there guarantees
    that they will exist, and remain constant, over time; b) are there official
    archival versions of the complete dataset (i.e., including the external
    resources as they existed at the time the dataset was created); c) are
    there any restrictions (e.g., licenses, fees) associated with any of the
    external resources that might apply to a future user? Please provide
    descriptions of all external resources and any restrictions associated with
    them, as well as links or other access points, as appropriate.
    } \\
    There are no fees. The Smithsonian data is hosted on a governmental website, which we believe will be well supported over time. The Sketchfab and Polycam data will be available to download easily on our platform. For Thingiverse and GitHub, the platforms are relatively stable and we expect most of the content to remain in place. For Thingiverse, an API key must be provided to download the content from their API. For GitHub, the data can be easily cloned. Users must follow the license of the content with which the original files were distributed and the terms of service for each platform.
    \\
    
    \textcolor{\sectioncolor}{\textbf{
    Does the dataset contain data that might be considered confidential (e.g.,
    data that is protected by legal privilege or by doctor-patient
    confidentiality, data that includes the content of individuals’ non-public
    communications)?
    }
    If so, please provide a description.
    } \\
    While rare, it is possible that confidential data exists as part of the 3D objects on the platforms.
    \\
    
    \textcolor{\sectioncolor}{\textbf{
    Does the dataset contain data that, if viewed directly, might be offensive,
    insulting, threatening, or might otherwise cause anxiety?
    }
    If so, please describe why.
    } \\
    While rare, it is possible that data that is considered offensive, insulting, threatening, or might cause anxiety exists as part of the 3D objects on the platforms.
    \\
    
    \textcolor{\sectioncolor}{\textbf{
    Does the dataset relate to people?
    }
    If not, you may skip the remaining questions in this section.
    } \\
    People may be present in the dataset, but only make up a small part of it. Section~\label{sec:analysis} discusses the results and analysis of running a face detector on renders of the objects. Most often, faces appear from dolls, historic sculptures, and anthropomorphic animations. Moreover, even including such data, only about 2.5\% captured faces.
     \\
    
    \textcolor{\sectioncolor}{\textbf{
    Does the dataset identify any subpopulations (e.g., by age, gender)?
    }
    If so, please describe how these subpopulations are identified and
    provide a description of their respective distributions within the dataset.
    } \\
    We do not identify people by subpopulation.\\
    
    \textcolor{\sectioncolor}{\textbf{
    Is it possible to identify individuals (i.e., one or more natural persons),
    either directly or indirectly (i.e., in combination with other data) from
    the dataset?
    }
    If so, please describe how.
    } \\
    If a scanned person is included in the dataset, it may be possible to visually identify them or identify them if their name is included as part of the metadata.
    \\
    
    \textcolor{\sectioncolor}{\textbf{
    Does the dataset contain data that might be considered sensitive in any way
    (e.g., data that reveals racial or ethnic origins, sexual orientations,
    religious beliefs, political opinions or union memberships, or locations;
    financial or health data; biometric or genetic data; forms of government
    identification, such as social security numbers; criminal history)?
    }
    If so, please provide a description.
    } \\
    While rare, it is possible that sensitive data may exist as part of the 3D objects on the platforms.
    \\
    
    \textcolor{\sectioncolor}{\textbf{
    Any other comments?
    }} \\
    No.\\

\subsection{Collection}

    \textcolor{\sectioncolor}{\textbf{
    How was the data associated with each instance acquired?
    }
    Was the data directly observable (e.g., raw text, movie ratings),
    reported by subjects (e.g., survey responses), or indirectly
    inferred/derived from other data (e.g., part-of-speech tags, model-based
    guesses for age or language)? If data was reported by subjects or
    indirectly inferred/derived from other data, was the data
    validated/verified? If so, please describe how.
    } \\
    The data was directly observable and hosted on several platforms, including GitHub, Thingiverse, Sketchfab, Polycam, and the Smithsonian Institute.
    \\
    
    \textcolor{\sectioncolor}{\textbf{
    Over what timeframe was the data collected?
    }
    Does this timeframe match the creation timeframe of the data associated
    with the instances (e.g., recent crawl of old news articles)? If not,
    please describe the timeframe in which the data associated with the
    instances was created. Finally, list when the dataset was first published.
    } \\
    Sketchfab data was collected as part of Objaverse 1.0. The new data was collected in Q1 \& Q2 of 2023.\\
    
    \textcolor{\sectioncolor}{\textbf{
    What mechanisms or procedures were used to collect the data (e.g., hardware
    apparatus or sensor, manual human curation, software program, software
    API)?
    }
    How were these mechanisms or procedures validated?
    } \\
    Python scripts were used to collect the data.\\
    
    \textcolor{\sectioncolor}{\textbf{
    What was the resource cost of collecting the data?
    }
    (e.g. what were the required computational resources, and the associated
    financial costs)
    } \\
    The cost of collecting the dataset was on the order of several thousand dollars, including costs to find, index, and download it. Resources included AWS CPU instances.
    \\
    
    \textcolor{\sectioncolor}{\textbf{
    If the dataset is a sample from a larger set, what was the sampling
    strategy (e.g., deterministic, probabilistic with specific sampling
    probabilities)?
    }
    } \\
    The dataset is filtered down based on licensing restrictions, duplicate content, and based on if the 3D object can successfully be imported into Blender.
    \\
    
    \textcolor{\sectioncolor}{\textbf{
    Who was involved in the data collection process (e.g., students,
    crowdworkers, contractors) and how were they compensated (e.g., how much
    were crowdworkers paid)?
    }
    } \\
    The data collection process was primarily performed by employed researchers at the Allen Institute for AI.
    \\
    
    \textcolor{\sectioncolor}{\textbf{
    Were any ethical review processes conducted (e.g., by an institutional
    review board)?
    }
    If so, please provide a description of these review processes, including
    the outcomes, as well as a link or other access point to any supporting
    documentation.
    } \\
    Institutional review boards were not involved in the collection of the dataset.
    \\
    
    \textcolor{\sectioncolor}{\textbf{
    Does the dataset relate to people?
    }
    If not, you may skip the remainder of the questions in this section.
    } \\
    People may be present in the dataset, but only make up a small part of it. Section~\label{sec:analysis} discusses the results and analysis of running a face detector on renders of the objects. Most often, faces appear from dolls, historic sculptures, and anthropomorphic animations. Moreover, even including such data, only about 2.5\% captured faces.
    \\
    
    \textcolor{\sectioncolor}{\textbf{
    Did you collect the data from the individuals in question directly, or
    obtain it via third parties or other sources (e.g., websites)?
    }
    } \\
    Data was collected from public facing platforms. On each of the platforms, users opted to make their data public.
    \\
    
    \textcolor{\sectioncolor}{\textbf{
    Were the individuals in question notified about the data collection?
    }
    If so, please describe (or show with screenshots or other information) how
    notice was provided, and provide a link or other access point to, or
    otherwise reproduce, the exact language of the notification itself.
    } \\
    Individuals were not notified about the collection of the dataset.
    \\
    
    \textcolor{\sectioncolor}{\textbf{
    Did the individuals in question consent to the collection and use of their
    data?
    }
    If so, please describe (or show with screenshots or other information) how
    consent was requested and provided, and provide a link or other access
    point to, or otherwise reproduce, the exact language to which the
    individuals consented.
    } \\
    Individuals were not notified about the collection of the dataset. \\
    
    \textcolor{\sectioncolor}{\textbf{
    If consent was obtained, were the consenting individuals provided with a
    mechanism to revoke their consent in the future or for certain uses?
    }
     If so, please provide a description, as well as a link or other access
     point to the mechanism (if appropriate)
    } \\
    Individuals were not notified about the collection of the dataset. \\
    
    \textcolor{\sectioncolor}{\textbf{
    Has an analysis of the potential impact of the dataset and its use on data
    subjects (e.g., a data protection impact analysis) been conducted?
    }
    If so, please provide a description of this analysis, including the
    outcomes, as well as a link or other access point to any supporting
    documentation.
    } \\
    An analysis has not been conducted.\\
    
    \textcolor{\sectioncolor}{\textbf{
    Any other comments?
    }} \\
    No.\\

\subsection{Processing / Cleaning / Labeling}

    \textcolor{\sectioncolor}{\textbf{
    Was any preprocessing/cleaning/labeling of the data
    done(e.g., discretization or bucketing, tokenization, part-of-speech
    tagging, SIFT feature extraction, removal of instances, processing of
    missing values)?
    }
    If so, please provide a description. If not, you may skip the remainder of
    the questions in this section.
    } \\
    Preprocessing the data was performed by computing renders of the objects and performing deduplication.
    \\

    \textcolor{\sectioncolor}{\textbf{
    Was the ``raw'' data saved in addition to the preprocessed/cleaned/labeled
    data (e.g., to support unanticipated future uses)?
    }
    If so, please provide a link or other access point to the “raw” data.
    } \\
s    The data that is downloaded contains the raw data and does not modify the individual files.\\

    \textcolor{\sectioncolor}{\textbf{
    Is the software used to preprocess/clean/label the instances available?
    }
    If so, please provide a link or other access point.
    } \\
    The software for cleaning the dataset and rendering will be made available.\\

    \textcolor{\sectioncolor}{\textbf{
    Any other comments?
    }} \\
    No.\\

\subsection{Uses}

    \textcolor{\sectioncolor}{\textbf{
    Has the dataset been used for any tasks already?
    }
    If so, please provide a description.
    } \\
    Yes, please see Section~\ref{sec:experiments} of the paper.\\

    \textcolor{\sectioncolor}{\textbf{
    Is there a repository that links to any or all papers or systems that use the dataset?
    }
    If so, please provide a link or other access point.
    } \\
    We recommend checking the Semantic Scholar page for the Objaverse-XL and Objaverse 1.0 papers to find up to date papers that use the dataset.
    \\

    \textcolor{\sectioncolor}{\textbf{
    What (other) tasks could the dataset be used for?
    }
    } \\
    The dataset could be used for a large number of use cases. Examples include making 3D tools more accessible (e.g., 3D impainting, text to 3D, image to 3D), robotic simulation \& embodied AI, training video models on animations, 2D vision tasks such as segmentation, and more.
    \\

    \textcolor{\sectioncolor}{\textbf{
    Is there anything about the composition of the dataset or the way it was
    collected and preprocessed/cleaned/labeled that might impact future uses?
    }
    For example, is there anything that a future user might need to know to
    avoid uses that could result in unfair treatment of individuals or groups
    (e.g., stereotyping, quality of service issues) or other undesirable harms
    (e.g., financial harms, legal risks) If so, please provide a description.
    Is there anything a future user could do to mitigate these undesirable
    harms?
    } \\
    Users should follow the license of the individual objects distributed as part of this dataset.
    \\

    \textcolor{\sectioncolor}{\textbf{
    Are there tasks for which the dataset should not be used?
    }
    If so, please provide a description.
    } \\
    New tasks must make sure to follow the license of the dataset and the license of the individual objects distributed as part of the dataset.
    \\

    \textcolor{\sectioncolor}{\textbf{
    Any other comments?
    }} \\
    No. \\

\subsection{Distribution}

    \textcolor{\sectioncolor}{\textbf{
    Will the dataset be distributed to third parties outside of the entity
    (e.g., company, institution, organization) on behalf of which the dataset
    was created?
    }
    If so, please provide a description.
    } \\
    Yes. The dataset will be made public.
    \\

    \textcolor{\sectioncolor}{\textbf{
    How will the dataset will be distributed (e.g., tarball on website, API,
    GitHub)?
    }
    Does the dataset have a digital object identifier (DOI)?
    } \\
    The dataset will be distributed through a Python API.
    \\

    \textcolor{\sectioncolor}{\textbf{
    When will the dataset be distributed?
    }
    } \\
    The dataset will be made publicly available towards the end of June, 2023.\\

    \textcolor{\sectioncolor}{\textbf{
    Will the dataset be distributed under a copyright or other intellectual
    property (IP) license, and/or under applicable terms of use (ToU)?
    }
    If so, please describe this license and/or ToU, and provide a link or other
    access point to, or otherwise reproduce, any relevant licensing terms or
    ToU, as well as any fees associated with these restrictions.
    } \\
    The dataset as a whole will be distributed under the ODC-By 1.0 license. The individual objects are subject to the licenses that they are released under, and users need to assess license questions based on downstream use.
    \\

    \textcolor{\sectioncolor}{\textbf{
    Have any third parties imposed IP-based or other restrictions on the data
    associated with the instances?
    }
    If so, please describe these restrictions, and provide a link or other
    access point to, or otherwise reproduce, any relevant licensing terms, as
    well as any fees associated with these restrictions.
    } \\
    The individual objects are subject to the licenses that they are released under, and users need to assess license questions based on downstream use. \\

    \textcolor{\sectioncolor}{\textbf{
    Do any export controls or other regulatory restrictions apply to the
    dataset or to individual instances?
    }
    If so, please describe these restrictions, and provide a link or other
    access point to, or otherwise reproduce, any supporting documentation.
    } \\
    No. \\

    \textcolor{\sectioncolor}{\textbf{
    Any other comments?
    }} \\
    No. \\

\subsection{Maintenance}

    \textcolor{\sectioncolor}{\textbf{
    Who is supporting/hosting/maintaining the dataset?
    }
    } \\
    The dataset will be hosted on Hugging Face.\\

    \textcolor{\sectioncolor}{\textbf{
    How can the owner/curator/manager of the dataset be contacted (e.g., email address)?
    }
    } \\
    Please contact \texttt{mattd@allenai.org}.\\

    \textcolor{\sectioncolor}{\textbf{
    Is there an erratum?
    }
    If so, please provide a link or other access point.
    } \\
    No.\\

    \textcolor{\sectioncolor}{\textbf{
    Will the dataset be updated (e.g., to correct labeling errors, add new
    instances, delete instances)?
    }
    If so, please describe how often, by whom, and how updates will be
    communicated to users (e.g., mailing list, GitHub)?
    } \\
    The dataset is currently self contained without immediate plans for updates.
    \\

    \textcolor{\sectioncolor}{\textbf{
    If the dataset relates to people, are there applicable limits on the
    retention of the data associated with the instances (e.g., were individuals
    in question told that their data would be retained for a fixed period of
    time and then deleted)?
    }
    If so, please describe these limits and explain how they will be enforced.
    } \\
    People may contact us to add specific samples to a blacklist.
    \\

    \textcolor{\sectioncolor}{\textbf{
    Will older versions of the dataset continue to be
    supported/hosted/maintained?
    }
    If so, please describe how. If not, please describe how its obsolescence
    will be communicated to users.
    } \\
    Objaverse 1.0 will continue to be supported.
    \\

    \textcolor{\sectioncolor}{\textbf{
    If others want to extend/augment/build on/contribute to the dataset, is
    there a mechanism for them to do so?
    }
    If so, please provide a description. Will these contributions be
    validated/verified? If so, please describe how. If not, why not? Is there a
    process for communicating/distributing these contributions to other users?
    If so, please provide a description.
    } \\
    We encourage others to build and extend the dataset for different use cases and may highlight some of those use cases if applicable.
    \\

    \textcolor{\sectioncolor}{\textbf{
    Any other comments?
    }} \\
    No

\section{Aesthetic Annotations}

\begin{figure}[b!]
    \includegraphics[width=\textwidth]{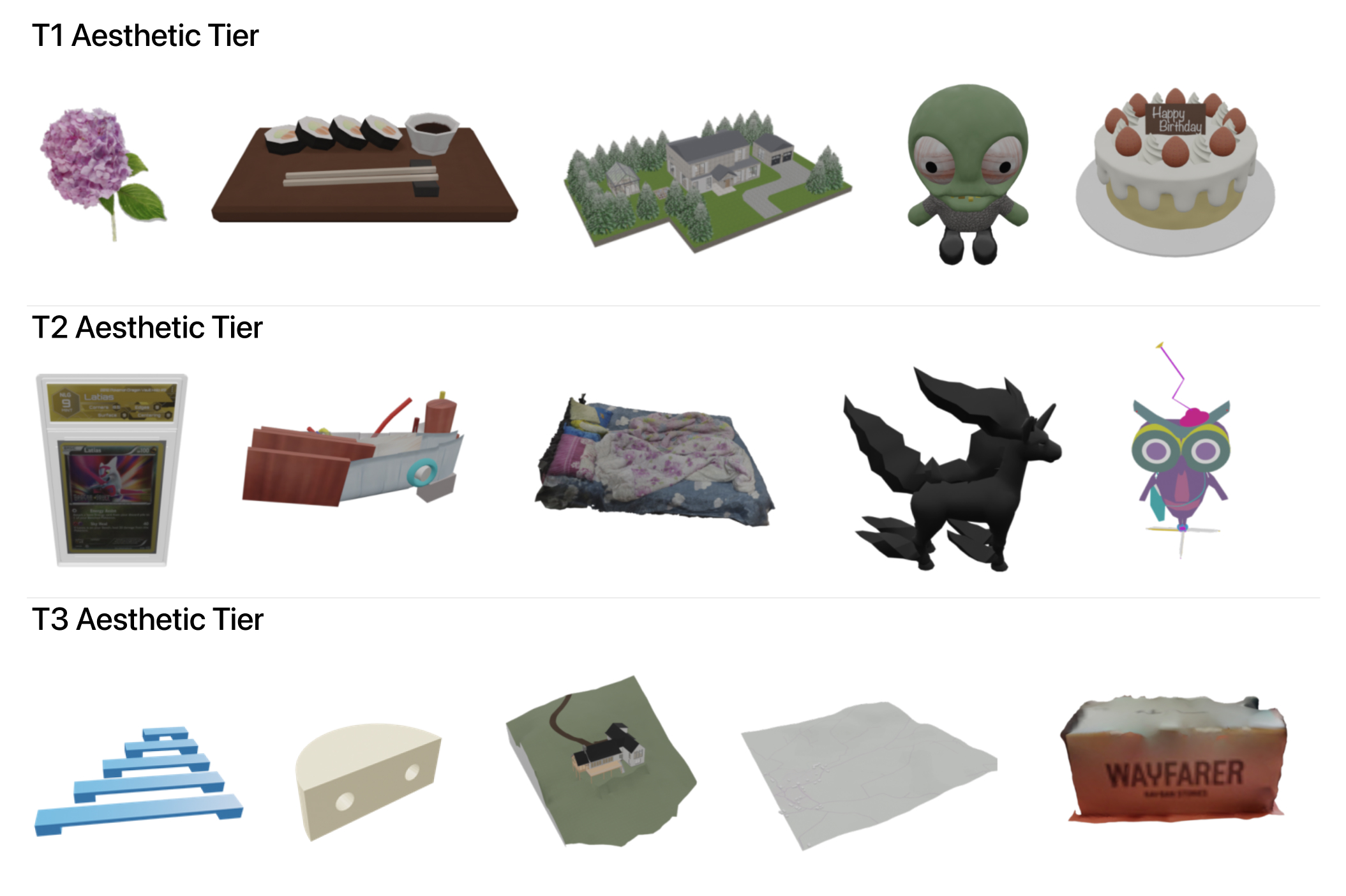}
    \caption{Random samples of renders showing LAION-Aesthetic V2 annotations across different tiers. Empirically, T1 tends to have the highest quality objects, followed by T2 and then T3.}
    \vspace{0.25in}
    \label{fig:aesthetic}
\end{figure}

We run LAION-Aesthetics V2~\cite{schuhmann2022laion} on renders of the objects, which can be used for filtering a higher quality subset of the objects. We group the objects into 3 tiers, which are depicted in Table~\ref{table:dataset_composition}. Figure~\ref{fig:aesthetic} shows examples of renders of objects placed on the different tiers.

\begin{table}[h!]
\centering
\begin{tabular}{cccc}
\toprule
\textbf{Category} & \textbf{Description} & \textbf{Aesthetic Score Cutoff} & \textbf{Percentage of Dataset} \\
\midrule
T1 & Highest aesthetic ranked objects & Greater than 4.5 & 14.2\% \\
T2 & Medium aesthetic ranked objects & Between 4 and 4.5 & 69.2\% \\
T3 & Low aesthetic ranked objects & Less than 4 & 16.6\% \\
\bottomrule
\end{tabular}
\vspace{0.025in}
\caption{LAION-Aesthetics V2 categorization for renders of Objaverse-XL objects.}
\label{table:dataset_composition}
\end{table}

\appendix

\end{document}